\documentclass[journal]{IEEEtai}

\usepackage{blindtext}
\usepackage{times}
\usepackage{epsfig}
\usepackage{amssymb}
\usepackage{amsthm}
\usepackage{graphicx}
\usepackage{amsfonts}
\usepackage{amsmath}
\usepackage{array}
\usepackage{mdwmath}
\usepackage{mdwtab}
\usepackage{eqparbox}
\usepackage{fixltx2e}
\usepackage{stfloats}
\usepackage{url}
\usepackage{diagbox}
\usepackage{verbatim}
\usepackage{booktabs}
\usepackage{multirow}
\usepackage{bm}
\usepackage{mathtools}
\usepackage{breqn}
\usepackage{float}
\usepackage{tcolorbox}
\usepackage{todonotes}
\usepackage{pgfmath}
\usepackage{siunitx}
\usepackage{adjustbox}
\usepackage{threeparttable}
\usepackage{hyperref}
\usepackage{flushend}

\usepackage[font=footnotesize,labelfont=bf]{caption}
\usepackage{subcaption}

\def\ourmethod{MoSegNAS}
\def\ourmodel{MoSegNet}
\DeclareMathOperator*{\argmin}{arg\,min}

\def\Minimize{\mathop{\rm Minimize}\limits}
\def\Maximize{\mathop{\rm Maximize}\limits}
\def\st{\mathop{\rm subject\ to}}

\definecolor{Gray}{gray}{0.85}
\definecolor{LightGray}{gray}{0.9}

\newcommand{\revision}[1]{{\textcolor{black}{#1}}}

\usepackage{algorithmic}
\usepackage[ruled, vlined, linesnumbered]{algorithm2e}

\begin{document}

\title{MoSegNAS: Surrogate-assisted Multi-objective Neural Architecture Search for Real-time Semantic Segmentation}

\author{Zhichao Lu,
        Ran Cheng,
        Shihua Huang,
        Haoming Zhang,
        Changxiao Qiu
        and~Fan Yang
}



\maketitle

\begin{abstract}

The architectural advancements in deep neural \textcolor{black}{networks} have led to remarkable leap-forwards across a broad array of computer vision tasks. 
Instead of relying on human expertise, neural architecture search (NAS) has emerged as a promising avenue toward automating the design of architectures. 
While recent achievements in image classification have suggested opportunities, the promises of NAS have yet to be thoroughly assessed on more challenging tasks of semantic segmentation.
The main challenges of applying NAS to semantic segmentation arise from two aspects: i) high-resolution images to be processed; ii) additional requirement of real-time inference speed (i.e. real-time semantic segmentation) for applications such as autonomous driving.
To meet such challenges, we propose a surrogate-assisted multi-objective method in this paper. 
Through a series of customized prediction models, our method effectively transforms the original NAS task into an ordinary multi-objective optimization problem. 
Followed by a hierarchical pre-screening criterion for in-fill selection, our method progressively achieves a set of efficient architectures trading-off between segmentation accuracy and inference speed. 
Empirical evaluations on three benchmark datasets together with an application using Huawei Atlas 200 DK suggest that our method can identify architectures significantly outperforming existing state-of-the-art architectures designed both manually by human experts and automatically by other NAS methods. 
\end{abstract}


\IEEEpeerreviewmaketitle

\section{Introduction}
\IEEEPARstart{D}{eep} Convolutional Neural Networks (CNNs) have been proven effective across a wide-range of machine learning tasks, including object recognition~\cite{alexnet}, speech recognition~\cite{speech_recognition}, language processing~\cite{sutskever2014sequence}, decision making~\cite{alphago}, etc. While improvements in computing hardware and training techniques have certainly contributed, architectural advancements have been undeniably the core driving forces behind these successes. Early progress primarily relies on skilled practitioners and elaborated designs. In computer vision, this manual process has led to designs such as VGG~\cite{vgg}, ResNet~\cite{resnet}, DenseNet~\cite{densenet} for object classification, and Faster R-CNN~\cite{rcnn}, DeepLab~\cite{deeplab} for object detection and segmentation. 

More recently, there has been a surge of interest in automating the design of network architectures. It is well-recognized now that the manual design process is a computationally impractical endeavor with respect to the increasing application scenarios of CNNs \revision{and many other types of deep neural networks \cite{9383028}}. Notably, neural architecture search (NAS) has emerged as a trending research topic for both academia and industries, as automatically generated network models exceeding the performance of manually designed ones on large-scale image classification problems~\cite{imagenet,nasnet2018}. While the recent algorithmic development of NAS on object classification has established a promising starting point \revision{for computer vision tasks}, the potential of NAS has yet to be fully assessed on more challenging and demanding tasks, such as semantic segmentation~\cite{pascal-voc}. 

\begin{figure*}[t]
    \centering
    \begin{subfigure}{0.8\textwidth}
    \centering
    \includegraphics[width=\textwidth{}]{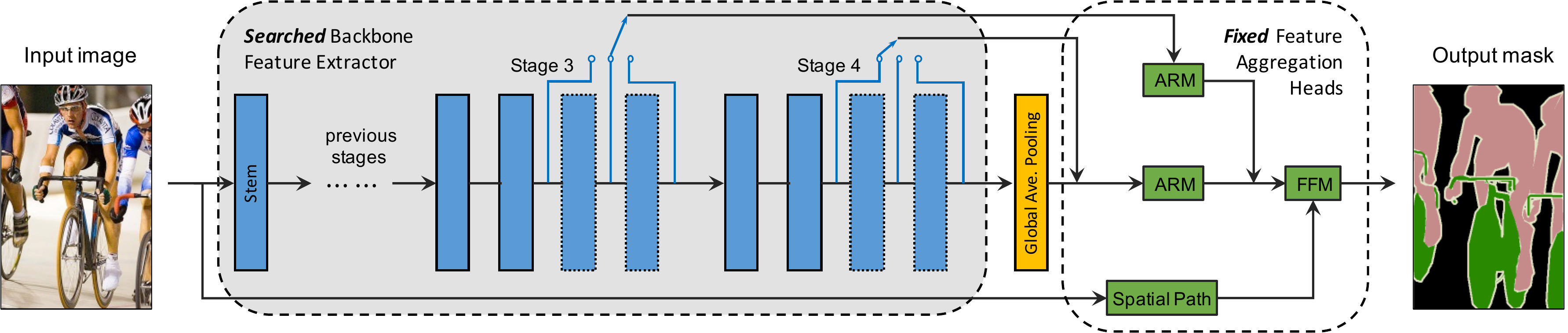}
    \end{subfigure}
\caption{\textbf{Search Space Overview:} The design of our search space follows the encoder-decoder convention. Each searchable architecture comprises a backbone CNN, as the encoder, for extracting features at multiple scales; and a decoder module to aggregate the extracted multi-scale features. Towards making the search tractable, we fix the decoder to the efficient segmentation heads from BiSeNet~\cite{yu2018bisenet} and focus on searching the architectural settings of the backbone CNN, including the input image scale, the No. of layers and output channels for each stage, and the No. of intermediate channels for each layer in the stage. \label{fig:search_space}}
\end{figure*}

The task of semantic segmentation performs classification at pixel-level--i.e., assigning a class label to every pixel in an input image. 
Thereby, in addition to the higher-level contextual information which is solely sufficient for object classification, semantic segmentation also requires rich spatial details to group pixels based on their semantic categories.
Modern deep learning based methods typically follow the process shown in Fig.~\ref{fig:search_space}. 
A CNN encoder (also referred as \emph{backbone}) is adopted to first extract features at multiple spatial resolutions, followed by another CNN decoder to aggregate these multi-scale features and output a prediction mask. 
The need of multi-scale representation and the necessity to operate on high resolution imagery result in steep computational requirement in evaluating the performance of an architecture for segmentation, hindering the utility of conventional simulation-based optimization means. 
Particularly, many important real-world scenarios of semantic segmentation impose rigorous speed requirements on CNN models to be practically useful, leading to the need of considering multiple objectives simultaneously. This special category is commonly referred as \emph{real-time} semantic segmentation \revision{(i.e., CNN models with inference speed of more than 30 images per second are typically considered to be real-time~\cite{yu2018bisenet,fan2021rethinking})}, with applications ranging from autonomous driving in navigation~\cite{cityscapes} to precision farming in agriculture~\cite{milioto2018real}.

Existing NAS methods for semantic segmentation either (i) opt for problem reformulation via continuous relaxation, which allows architecture parameters to be jointly optimized with the weights, to search the entire architecture all at once~\cite{liu2019auto,zhang2019customizable}; or (ii) only search the decoder architecture while inheriting a generic backbone CNN from object classification~\cite{nekrasov2019fast}.
Despite the efficiency provided from the continuous relaxation, the reformulated problem \textcolor{black}{tends} to bias architectures with faster convergence, leading to potentially sub-optimal architectures, as faster converging architectures need not necessarily be the ones that also generalize better~\cite{Shu2020Understanding}. 
Moreover, incorporating additional application-related objectives in continuous relaxation based methods, which rely on gradients to update architectural parameters, is not straightforward. 

Concurrently, evolutionary algorithms (EAs) have earned a plethora of attention in NAS~\cite{real2019regularized,8742788,nsganetv1,8712430,9360872}. 
In an EA, a set of solutions are processed in parallel to approximate the optimal architectures, where the selection is carried out on the relative differences among solutions, obviating the need of gradient estimation from continuous relaxation. 
Despite that the population-based nature of EAs enables a flexible extension to handle multiple objectives, direct porting ideas from  the current literature of EAs would not suffice the requirements of NAS for real-time semantic segmentation -- the computational cost caused by the performance evaluations of architectures obtained in every generation can be prohibitively expensive.  

In this paper, we propose \emph{\ourmethod{}} to breach this steep computational barrier and present the first attempt of tailoring evolutionary multi-objective optimization based NAS for real-time semantic segmentation, with the assistance of surrogate modeling. 
Through a series of customized prediction models, \ourmethod{} effectively transforms the NAS task to an ordinary multi-objective optimization problem, such that existing multi-objective evolutionary algorithms can be applied to obtaining solutions (i.e. architectures) trading-off between segmentation accuracy and inference speed. 
A partial set from the obtained solutions is selected via a hierarchical pre-screening criterion, and then used to refine the learned prediction models. 
These two steps are alternated in iterations until a computation budget is exhausted. The key contributions are summarized below:

\vspace{2pt}
\begin{itemize}
    \item We introduce \ourmethod{} as an alternative to existing NAS methods for semantic segmentation. 
    Instead of replying on estimated gradients from continuous relaxation, we advocate for a surrogate-assisted evolutionary multi-objective framework. 
    \item \ourmethod{} adopts an online surrogate model for predicting the segmentation accuracy. With a sequence of modifications introduced to a multi-layer perceptron, we demonstrate that an indicative predictor can be efficiently learned with few hundreds of samples. 
    \item \ourmethod{} adopts an offline surrogate model, in the form of a look-up table, for predicting inference latency. 
    Considering the fact that high-fidelity evaluation of latency is an-order-of-magnitude cheaper than segmentation accuracy in terms of simulation time, \ourmethod{} is further equipped with a customized hierarchical pre-screening criterion for in-fill selection.    
    \item We demonstrate that \ourmethod{} leads to state-of-the-art performance on real-time semantic segmentation, achieving higher accuracy and inference speed on Cityscapes dataset~\cite{cityscapes} than architectures designed both manually and by other NAS methods. 
    Under transfer learning setup, we demonstrate that the obtained models from \ourmethod{} also lead to state-of-the-art performance on COCO-Stuff-10$K$~\cite{caesar2018coco} and PASCAL VOC 2012~\cite{pascal-voc} datasets.
    \item We demonstrate the practical utility of \ourmethod{} in designing hardware-dependent models on an application using Huawei Atlas 200 DK, where the obtained models consistently outperform existing models across a spectrum of inference latency. 
\end{itemize}

The remainder of the paper is organized as follows. Section~\ref{sec:related} summarizes the related literature. The proposed method is elaborated in Section~\ref{sec:approach}. In Section~\ref{sec:exper}, we describe the experimental setup to validate our method along with discussions of the results, followed by ablative experiments and an application study in Sections~\ref{sec:ablation}.
Section~\ref{sec:application} presents the results of applying our method to Huawei Atlas 200 Developer Kit. Finally, we conclude with a brief summary of the proposed method and our findings in Section~\ref{sec:conclusion}.
 \section{Related Work\label{sec:related}}
NAS has been overwhelmingly successful in the recent past for automatically customizing network models that are tailored to the task at hand. 
The early development of NAS primarily concentrated on the fundamental task in computer vision -- i.e., image classification. 
Relying on reinforcement learning (RL), continuous relaxation (i.e. gradient), or evolutionary algorithms (EAs), impressive empirical results surpassing human-level performance have been reported on various image classification benchmarks~\cite{imagenet}. 
Moving beyond classification, many recent attempts aiming to adapt NAS methods to higher-level tasks have been reported. 
Chen el at.~\cite{chen2018searching} presented the first effort of applying the RL-based NAS method proposed in~\cite{nasnet2018} to the domain of dense image prediction.
Similarly, NAS-FPN~\cite{ghiasi2019fpn} learned a recurrent neural network (RNN) as a controller to automatically design the architectures of feature pyramid networks for object detection.
DetNAS~\cite{chen2019detnas} used NAS to demonstrate the importance of backbone architecture design for object detection.
Auto-DeepLab proposed a hierarchical search space that allows both macro and micro structures of a network to be jointly optimized, achieving state-of-the-art results on semantic segmentation~\cite{liu2019auto}. 
In the remainder of this section, we provide a brief overview on methods closely related to the technical aspects of this paper. 

\vspace{3pt}
\noindent\textbf{Surrogate Modelling:} The main computation bottleneck of NAS resides in the performance evaluations of architectures. 
Many surrogate-based methods have been proposed to expedite the evaluation of architectures for image classification. 
In general, existing methods can be broadly classified into two categories. 
The first category focuses on mitigating the computation cost of a single high-fidelity evaluation by reducing the number of training epochs that are required before the performance of an architecture can be assessed. 
Common methods include i) weight sharing that allows offspring architectures to inherit weights from parents or a super network as a warm-start~\cite{one-shot,liu2018darts,nsganetv2}; 
ii) early stop that terminates the training process at an earlier phase by either heuristics~\cite{nasnet2018,mnasnet} or extrapolation~\cite{baker2017metaqnn}; 
iii) partial training samples that lead to reduction in epochs required for training to converge~\cite{8712430}. 

The second category focuses on reducing the number of high-fidelity evaluations required to drive their algorithms towards optimal architectures. 
Methods in this group mostly resort to learning accuracy predictors~\cite{9336721}. 
E2EPP~\cite{e2epp} adopted an off-line random forest based surrogate model to directly predict performance from architectural encoding.  
OnceForAll~\cite{onceforall} used \textcolor{black}{an} MLP to predict accuracy, where the surrogate model is also learned in an offline manner, thus requiring a large number of training samples.
ChamNet~\cite{chamnet} improved the sample efficiency of offline surrogate modeling by selectively sampling architectures with diverse complexity (FLOPs, latency, energy, etc.) to construct the accuracy predictor. 
Further improvement introduced in NSGANetV2~\cite{nsganetv2} adopted online surrogate modelling that used architecture search to guide the construction of the accuracy predictor, thus substantially reducing the number of training samples. 

Despite the impressive advancement, the surrogate models adopted by most existing methods are generic or even simplistic. 
In contrast, \ourmethod{} applied a series of enhancements (sparse encoding, ranking loss, synthetic data, etc.) to a conventional MLP model, demonstrating the first attempt to predict the segmentation performance of neural architectures using surrogate models. 

\vspace{3pt}
\noindent\textbf{Multi-objective NAS:}
A plethora of hardware-dependent NAS works have been introduced lately for image classification~\cite{cai2018proxylessnas,mnasnet,wan2020fbnetv2}. 
These methods sought to improve models' computational efficiency (i.e. inference latency, power consumption, memory footprint) on a specific hardware while trading-off classification accuracy to a small extent. 
A common theme behind these methods is to adopt a scalarized objective function or an additional constraint to encourage high accuracy and penalize compute inefficiency at the same time. 
CAS~\cite{zhang2019customizable} studied various resource-related constraints and demonstrated the utility of NAS on real-time semantic segmentation. 
It applied continuous relaxation to an extended cell-based search space~\cite{liu2018darts}, allowing the architectural parameters to be optimized jointly with weights via stochastic gradient descent. 
As a follow-up work, FasterSeg~\cite{Chen2020FasterSeg} further reduced the run-time latency of the searched models by incorporating prior wisdom (e.g. multi-scale branches~\cite{hrnetv2} and dilated convolution~\cite{chen2018encoder}) to the search space. 

Conceptually, the search of architectures (in the aforementioned works) is still guided by a single objective and thus only one architecture is obtained per search. 
Empirically, multiple runs with different weighting of the objectives are required to find an architecture with the desired trade-off, or multiple architectures with different complexities. 
Concurrently, another streamlining of multi-objective NAS works for image classification emerged, aiming to approximate the entire Pareto-optimal front simultaneously in a single run~\cite{jin2008pareto,nsganet,elsken2018efficient,9268174,nat}. 
These methods utilize heuristics and relative differences among individuals in a population to efficiently navigate through the search space, allowing practitioners to visualize and choose a suitable model \emph{a posteriori} to the search. 
Our proposed \ourmethod{}, belonging to this category, is the first tailored multi-objective NAS method for real-time semantic segmentation. 

\begin{algorithm}[t]
\SetAlgoLined
\SetKwInOut{Input}{Input}
\SetKwInOut{Output}{Output}
\SetKwFor{For}{for}{do}{end for}
\footnotesize
\Input{Training data $\mathcal{D}_{trn}$, validation data $\mathcal{D}_{vld}$, initial population size $N$, max. \# of generations $T$, \# of high-fidelity eval. per gen. $K$.}
    $t$ $\leftarrow$ 0 \textcolor{gray}{// initialize an iteration counter}.\\
    $\mathcal{A}$ $\leftarrow$ $\emptyset$ \textcolor{gray}{// initialize an archive to store all evaluated architectures}.\\
    $\mathcal{H}$ $\leftarrow$ train a surrogate model, \emph{hypernetwork}, to generate weights for candidate architectures (see Sec.~\ref{sec:psi}). \\
    ${L}u{T}$ $\leftarrow$ construct a surrogate model, \emph{look-up table}, to predict latency (see Sec.~\ref{sec:lut}). \\
    \textcolor{gray}{// Initialize the parent population (see Sec.~ in the supp. materials).} \\
    $P$, $F_{p}$, $\mathcal{A}$ $\leftarrow$ $Initialization$($N$, $\mathcal{H}$, $\mathcal{D}_{vld}$, $\mathcal{A}$) \\
    \While{$t < T$}{
        \textcolor{gray}{// Construct a surrogate model to predict accuracy.} \\
        $\tilde{f}$ $\leftarrow$ $ConsAccPred$($\mathcal{A}$) \hspace{29mm} $\triangleleft$ Algo.~\ref{algo:ranknet}\\
        \textcolor{gray}{// Generate offspring population using a MOEA (e.g., NSGA-II~\cite{nsga2}, MOEA/D~\cite{moead}) with surrogate objectives.} \\
        $Q$ $\leftarrow$ MOEA($\tilde{f}$, ${L}u{T}$) \\
        \textcolor{gray}{// Select top-$K$ offspring for high-fidelity evaluation} \\
        $Q^*$ $\leftarrow$ $PreScreen$($Q$, $P$, $K$) \hspace{23mm} $\triangleleft$ Algo~\ref{algo:pre-screen}\\
        \textcolor{gray}{// Evaluate the selected offspring.} \\
        $F_{q}$ $\leftarrow$ $Evaluate$($Q^*$, $\mathcal{H}$, $\mathcal{D}_{vld}$) \hspace{21mm} $\triangleleft$ Algo~\ref{algo:eval}\\
        $\mathcal{A}$ $\leftarrow$ $\mathcal{A}$ $\cup$ ($Q^*$, $F_p$) \textcolor{gray}{// archive the evaluated offspring.} \\
        $t$ $\leftarrow$ $t + 1$ \textcolor{gray}{// repeat above steps for $T$ generations}. \\
    }
    $P^*$ $\leftarrow$ choose top-ranked architectures based on trade-offs (see Sec.~1 in  supp. materials).\\
\textbf{Return} $P^*$
\caption{\small General framework of \ourmethod{} \label{algo:framework}}
\end{algorithm}

\section{Proposed Method\label{sec:approach}}
As summarized in Algorithm~\ref{algo:framework}, \ourmethod{} employs a series of surrogate modeling techniques to expedite the search of architectures. 
First, it learns a surrogate model, in the form of a hypernetwork, to approximate the reaction set mapping, through which the inner optimization loop over the network weights is obviated. 
Then, in each iteration of the search, an accuracy predictor is learned from previously evaluated architectures to rank newly generated offspring. 
Jointly with the latency look-up table (i.e. a learned surrogate model to predict the inference speed of a network), an auxiliary problem is constructed. 
Afterward, the optimization outcomes from a standard multi-objective evolutionary algorithm (MOEA) become the candidate architectures, from which a subset is selected for (high-fidelity) evaluation and archived. 
At the end of this evolutionary process, architectures with the best trade-offs will be selected as outputs. 
In the remainder of this section, we first formally define the problem and introduce the encoding strategy in Section~\ref{sec:formulation} and \ref{sec:encoding} respectively, followed by detailed descriptions on each component of \ourmethod{} in Sections~\ref{sec:psi}--\ref{sec:pre-screen}.

\subsection{Problem Formulation\label{sec:formulation}}
In this work, we approach the task of real-time semantic segmentation from a multi-objective perspective and mathematically formulate the problem as a bilevel optimization problem to simultaneously maximize segmentation accuracy and inference speed. 
For a targeted dataset $\mathcal{D} = \{\mathcal{D}_{trn}, \mathcal{D}_{vld}, \mathcal{D}_{tst}\}$, the problem is defined as:
\begin{equation}
\small
\begin{aligned}
\Maximize_{\bm{x}} & \hspace{3mm} \Big\{f_1\big(\bm{x}; \bm{w}^*\big),~f_2\big(\bm{x}\big)\Big\}, \\
\st  & \hspace{3mm} \bm{w}^* \in \argmin_{\bm{w}}~\mathcal{L}oss\big(\bm{w}; \bm{x}\big), \\
     & \hspace{3mm} \bm{x} \in \mathbf{\Omega}_{x}, \hspace{3mm} \bm{w} \in \mathbf{\Omega}_{w}.
\end{aligned}
\label{def:nas}
\normalsize
\end{equation}

\noindent where $\bm{\Omega}_{x} = \Pi_{i=1}^{n}[a_i, b_i] \subseteq \mathbb{Z}^n$ is the architecture decision space; $a_i$, $b_i$ are the lower and upper bounds, respectively. 
The variables $\bm{x} = (x_1, \ldots, x_n)^T \in \bm{\Omega}_x$ define a candidate architecture, and the variables $\bm{w} \in \bm{\Omega}_w \subseteq \mathbb{R}^m$ denote its associated weights. 
We use $f_1(\cdot)$, $f_2(\cdot)$ to denote the segmentation accuracy on the validation data $\mathcal{D}_{vld}$ and the inference speed on a specific hardware, respectively. $\mathcal{L}oss(\bm{w}; \bm{x})$ is the training loss of an architecture $\bm{x}$ on the training data $\mathcal{D}_{trn}$. 
We reserve $\mathcal{D}_{tst}$ for comparison with other methods.

The hierarchical nature of the problem presented in (\ref{def:nas}) necessitates a nested loop of optimizations, i.e., an inner optimization over the network weights $\bm{w}$ for a given architecture $\bm{x}$ and an outer optimization over the network architectural variables $\bm{x}$ themselves. \revision{It is worth noting that the bilevel optimization, on its own, is a challenge research topic \cite{sinha2017review,9766417}, and it is beyond the main scope of this paper.}

\subsection{Search Space and Encoding\label{sec:encoding}}

In line with prior wisdom, we follow the encoder-decoder architectural framework and leverage the BiSeNet~\cite{yu2018bisenet} heads as the decoder, and focus on designing the encoder. 
A pictorial illustration is provided in Fig.~\ref{fig:search_space}. 
We start with the premise of constructing a search space that may express most of the well-established backbone CNNs for segmentation tasks~\cite{rcnn}. 
The overall structure of our encoder network consists of a stem, four stages and a tail of global averaging pooling. 
The \emph{stem} and \emph{stages} are searched, while the \emph{tail} is handcrafted and common to all networks. 
Each \emph{stage} comprises of multiple layers, and each \emph{layer} follows the residual bottleneck structure, i.e., a sequence of a $1\times1$ convolution for compression, a $3\times3$ convolution, and another $1\times1$ convolution for expansion~\cite{resnet}. 

\ourmethod{} then searches over four important dimensions constituting the design of an encoder network, including the input image scale, the width (in terms of \# of channels), the depth (in terms of \# of layers), and the locations for outputting features. 
We encode these choices in an integer-valued string, and pad zeros to avoid variable-length representations (as the depth of each stage is also searched). 
With one digit for input scale, two digits for stem, six digits for each stage (except stage 3 where it uses eight digits), the total length of the encoding is 29. 
The resulting search space contains approximately \num{e14} unique architecture designs. It is worth mentioning that we use one-shot encoding to sparsify the integer string representation for learning the accuracy surrogate model, which is elaborated in the following subsections. 

Readers can refer to the supplementary materials for a more detailed break-down of our search space in Table~1 and an example of our encoding in Fig.~1. 

\subsection{Surrogate Modeling for Segmentation Accuracy\label{sec:acc_surrogate}}

\noindent\textbf{Single-level Reduction.\label{sec:psi}}
The main computational bottleneck of solving the problem in (\ref{def:nas}) stems from the fact that every evaluation of segmentation accuracy ($f_2$) invokes another optimization to be performed for learning the optimal weights ($\bm{w}^*$). 
Hence, as the first step to improve the efficiency of our approach, we aim to remove the inner loop of weights optimization via surrogate modelling. 
More specifically, let us consider an equivalent formulation of the problem, stated in terms of the reaction set mapping $\Psi: \mathbb{Z}^n \rightrightarrows \mathbb{R}^m$, as follows~\cite{sinha2017review}:
\begin{equation}
\small
\begin{aligned}
& \Psi(\bm{x}) = \argmin_{\bm{w} \in \mathbf{\Omega}_{w}}\mathcal{L}oss\big(\bm{w}; \bm{x}\big),
\end{aligned}
\label{def:spi}
\normalsize
\end{equation}

\noindent which represents the constraint defined by the inner optimization problem, i.e., $\Psi(\bm{x}) \subset \mathbf{\Omega}_{w}$ for every $\bm{x} \in \mathbf{\Omega}_{x}$. The nested loop of optimizations in (\ref{def:nas}) can then be re-formulated as a constrained single-level optimization problem, as below:

\begin{equation}
\small
\begin{aligned}
\Maximize_{\bm{x} \in \mathbf{\Omega}_{x},~\bm{w} \in \mathbf{\Omega}_{w}} & \hspace{3mm} \Big\{f_1\big(\bm{x}; \bm{w}^*\big),~f_2\big(\bm{x}\big)\Big\}, \\
\st  & \hspace{3mm} \bm{w}^* \in \Psi(\bm{x}).
\end{aligned}
\label{def:nas-psi}
\normalsize
\end{equation}

\noindent where $\Psi$ can be interpreted as a parameterized constraint for the variables $\bm{w}$ of the original inner optimization in (\ref{def:nas}). 

If the $\Psi$-mapping can somehow be determined, it effectively reduces a hierarchical problem to a regular single-level optimization, thereby, the costly iterations of the inner optimization are avoided. However, in the context of neural architecture search, the $\Psi$-mapping between architectures and their optimal weights is seldom defined analytically due to the non-linear nature of modern CNNs (e.g., non-linear activation and pooling functions). Thereby, we opt for the route of surrogate modelling the reaction set mapping in this work. 

\begin{figure}[!hbt]
    \centering
    \begin{subfigure}[b]{0.48\textwidth}
    \centering
    \includegraphics[width=\textwidth{}]{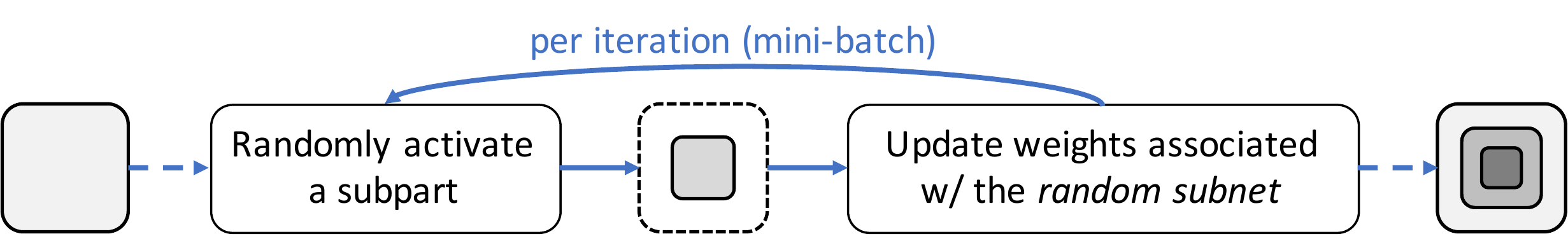}
    \end{subfigure} \\
    \centering
    \vspace{1em}
    \begin{subfigure}[b]{0.48\textwidth}
    \centering
    \includegraphics[width=.95\textwidth{}]{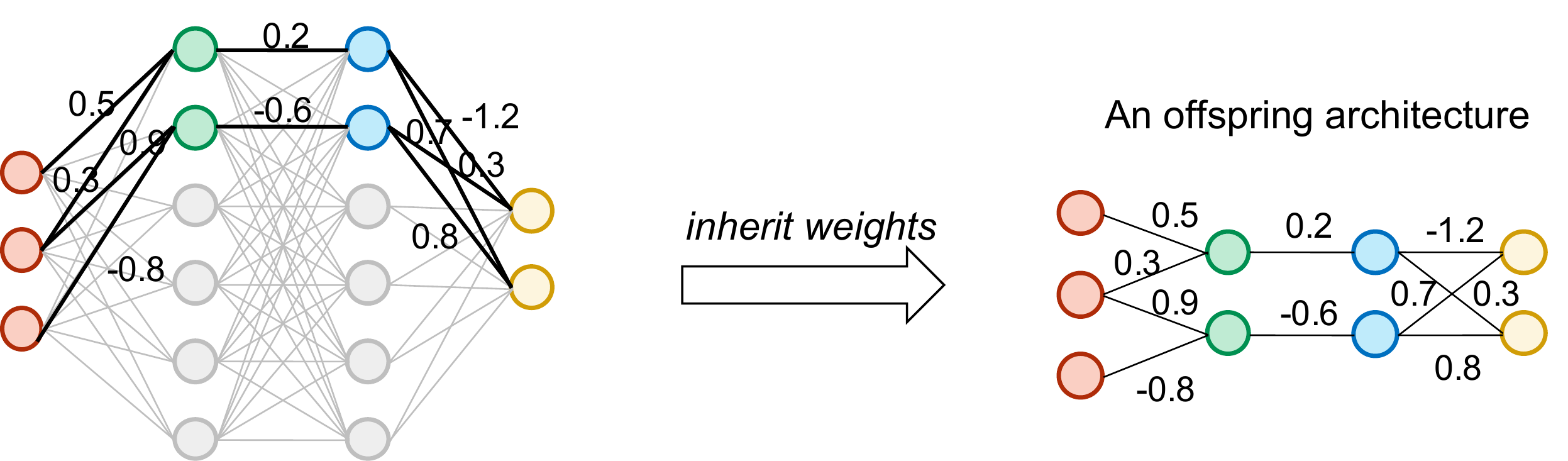}
    \end{subfigure}
\caption{\textbf{Top:} For learning a hypernetwork, a sub-part of the hypernetwork is randomly sampled during each iteration and only the associated parameters are updated via stochastic gradient descent. \textbf{Bottom:} With a trained hypernetwork, the segmentation accuracy of a candidate architecture is evaluated with the weights inherited from the hypernetwork. \label{fig:hypernetwork}}
\end{figure}

Following recent advances in one-shot and weight sharing NAS works~\cite{one-shot}, we model the $\Psi$-mapping in the form of a hypernetwork $\mathcal{H}(\theta; \bm{x})$, which is trained (by optimizing its parameters $\theta$) to generate weights conditioned on a given architecture~\cite{brock2018smash}. In \ourmethod{}, the $\mathcal{H}(\cdot)$ corresponds to the largest architecture\footnote{The largest architecture is defined by setting the searched architectural parameters to their upper bounds.} encoded in the search space, such that all searchable architectures become sub-parts. For learning the $\mathcal{H}(\cdot)$, a random sub-part ($\bm{x}_i$) is activated for forward passes during each iteration of training, and only the parameters associated with the activated sub-part ($\theta(\bm{x}_i)$) will be updated. The trained hypernetwork then becomes an approximation of the actual $\Psi$-mapping, from which the optimal weights ($\bm{w}^*$) of an architecture can be directly obtained. A pictorial illustration is provided in Fig~\ref{fig:hypernetwork}. And the evaluation of segmentation accuracy becomes an inference on the validation data, as shown in Algorithm~\ref{algo:eval}.

\begin{algorithm}[tbh]
\SetAlgoLined
\SetKwInOut{Input}{Input}
\SetKwInOut{Output}{Output}
\SetKwFor{For}{for}{do}{end for}
\footnotesize
\Input{Individuals $X$, hypernetwork $\mathcal{H}$, validation data $\mathcal{D}_{vld}$.}
    $F$ $\leftarrow$ $\emptyset$ \textcolor{gray}{// initialize a list to store objective vectors}.\\
    \For{$x$ in $X$}{
        \emph{net} $\leftarrow$ decode architecture $x$ to a neural network.\\
        $w$ $\leftarrow$ $\mathcal{H}$($x$) \textcolor{gray}{// inherit weights from the trained hypernetwork}.\\
        $t_o$ $\leftarrow$ time() \textcolor{gray}{// initialize a latency timer}. \\
        $acc$ $\leftarrow$ \emph{net}($w$, $\mathcal{D}_{vld}$) \textcolor{gray}{// evaluate segmentation accuracy on validation data using inherited weights directly}.\\
        $lat$ $\leftarrow$ (time() - $t_o$) / \emph{length}($\mathcal{D}_{vld}$) \textcolor{gray}{// measure latency}.\\
        $F$ $\leftarrow$ $F$ $\cup$ ($acc$, $lat$) \\
    }
\textbf{Return} $F$
\caption{\small Performance evaluation of architectures\label{algo:eval}}
\end{algorithm}

\begin{figure}[!hbt]
    \centering
    \begin{subfigure}{0.48\textwidth}
    \centering
    \includegraphics[width=.95\textwidth{}]{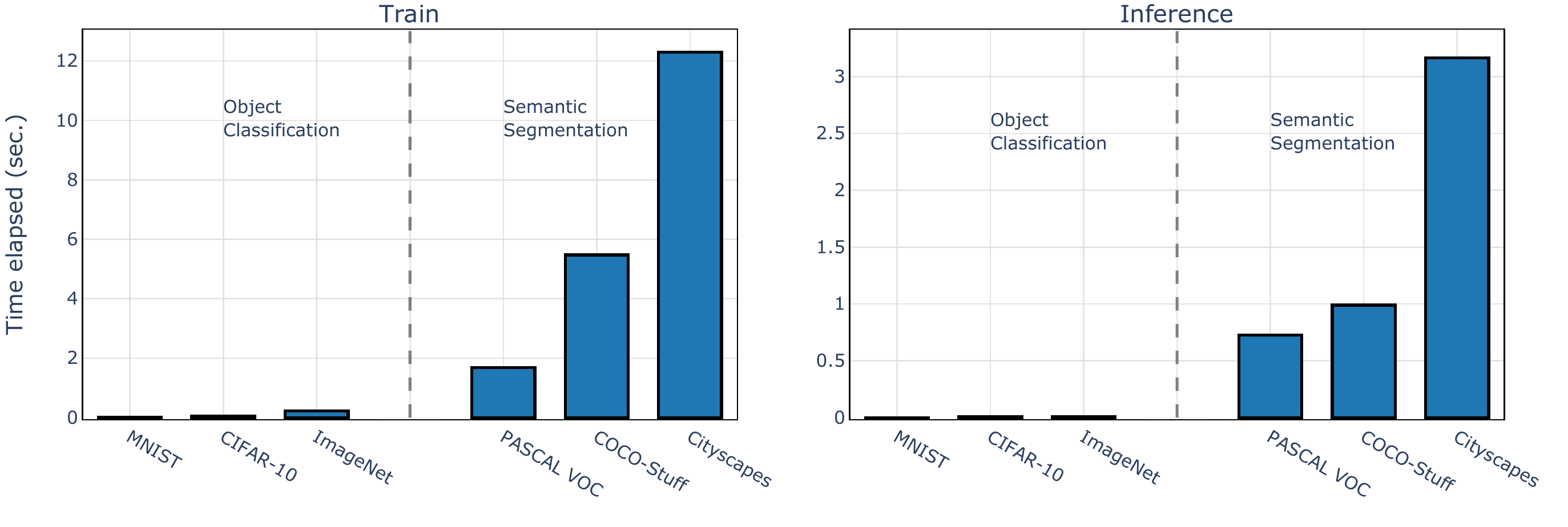}
    \end{subfigure}
\caption{\textbf{Wall-clock time complexity comparison} between object classification and semantic segmentation tasks. We record the time to process a batch of 64 images during training (\emph{Left}) and inference (\emph{Right}).\label{fig:speed}}
\end{figure}

\vspace{3pt}
\noindent\textbf{Segmentation Accuracy Prediction.\label{sec:ranknet}}
Recall that the task of semantic segmentation assigns a class label to every pixel in an input image, which requires intermediate features to be processed at high-resolution in a CNN~\cite{hrnetv2}. As a result, the cost of evaluating a network for semantic segmentation by forward inference is still considerably high due to a low batch size, as opposed to object classification (see Fig~\ref{fig:speed} for a visual comparison). Therefore, extensive inference by probing the approximated $\Psi$-mapping can still render the entire search computationally prohibitive. To mitigate this computational burden, we further develop a surrogate model to directly predict the segmentation accuracy of an architecture from its architectural parameters. By learning a functional relation between the architectural representations and the corresponding accuracy, our surrogate predictor disentangles the evaluation of an architecture from data-processing. Consequently, the cost of a single evaluation reduces from minutes to less than a second. 

In this work, we adopt a multi-layer perceptron (MLP). For the purpose of regression, a MLP is conventionally trained with the mean square error (MSE) loss that minimizes the Euclidean distances between ground truth and network output values, as below:

\begin{equation}
\begin{aligned}
\mathcal{L}_{mse} = \frac{1}{N}\sum_{i}~\lVert\hat{f}(\bm{x}_i; \textcolor{black}{\bm{\Theta}}) - y_{i}\rVert^2_2,
\end{aligned}
\label{def:reg_loss}
\normalsize
\end{equation}
where $\hat{y}_i = \hat{f}(\bm{x}_i; \textcolor{black}{\bm{\Theta}}) \in \mathbb{R}$ is the output of a neural network on the $i$th training sample $\bm{x}_i$, and ${y}_i \in \mathbb{R}$ is the ground truth label. $N$ is the number of training samples. However, for an evolutionary algorithm (as in our case) where the selections are carried out on the relative differences among solutions, a low MSE is desirable but \emph{not} necessary. In line with this derivation, we instead use a ranking loss to explicitly align the training of the MLP with the goal of maximizing the rank-order correlation in predictions, as follows:

\begin{equation}
\small
\begin{aligned}
\mathcal{L}_{rank} & = \frac{1}{2N}\sum_{i, j} max\big(0, \gamma - \delta(\hat{y}_i,~\hat{y}_j)({y}_i - {y}_j)\big), \\
\delta(\hat{y}_i, \hat{y}_j) & =
\begin{cases}
    1, & \text{if $\hat{y}_i > \hat{y}_j$,}\\
    -1, & \text{otherwise.}
\end{cases}\\
\end{aligned}
\label{def:rank_loss}
\normalsize
\end{equation}
where $\hat{y}_i = \hat{f}(\bm{x}_i; \textcolor{black}{\bm{\Theta}})$, $\hat{y}_j = \hat{f}(\bm{x}_j; \textcolor{black}{\bm{\Theta}})$ are the outputs of the MLP on the $i$th, $j$th training samples, respectively. While ${y}_i \in [0, 1]$, ${y}_j \in [0, 1]$ are the normalized ground truth labels for the $i$th, $j$th training samples, respectively. To be consistent, we also map $\hat{y}_i$ and $\hat{y}_j$ to $[0, 1]$ via a sigmoid function. $\delta(\hat{y}_i,~\hat{y}_j)$ is an utility function that indicates the relative ranking of the two inputs. And $\gamma$ is a hyperparameter controlling the margin. A similar concept can be found in the step of maximum-margin classification in a support vector machine. With the added $\ell_2$-norm of the weights to prevent over-fitting, the loss that we use to backpropagate for training the MLP is defined as:

\begin{equation}
\begin{aligned}
\mathcal{L}oss = \mathcal{L}_{rank} + \textcolor{black}{\beta} \lVert\bm{w}\rVert^2_2,
\end{aligned}
\label{def:loss}
\normalsize
\end{equation}
where \textcolor{black}{$\beta$} is the regularization parameter, which is set at a small value, e.g. \num{2.5e-4}. 

In addition to high rank-order correlation, another desired property of a surrogate model is \emph{sample efficiency}. It is well-known in the literature that neural networks generalize well only when there is sufficient amount of data. 
To overcome this barrier without excessive high-fidelity evaluations, we focus on generating and using synthetic data to complement the training of the MLP under limited genuine data. 
Inspired by the teacher-student knowledge distillation~\cite{hinton2015distilling}, we first train a set of widely-used surrogate models (e.g., Radial Basis Function (RBF), Decision Tree (DT), Gradient Boosting (GB)~\cite{ke2017lightgbm}, etc.) as teachers prior to the MLP training. 
In each iteration of the subsequent training, we randomly sample a small set of extra data with labels predicted by these teachers. 
We then augment the original data with these synthetic data and proceed to the stochastic gradient descent.
For reference, we name the proposed accuracy predictor as \emph{RankNet} and the pseudocode outlining the process of building RankNet is provided in Algorithm~\ref{algo:ranknet}. 

\begin{algorithm}[tbh]
\SetAlgoLined
\SetKwInOut{Input}{Input}
\SetKwInOut{Output}{Output}
\SetKwFor{For}{for}{do}{end for}
\footnotesize
\Input{Archive containing all past evaluated solutions $\mathcal{A}$, \\\# of epochs $T$, weight decay $\lambda$, initial learning rate $\eta_{o}$.}
    $\hat{X}$, $\hat{Y}$ $\leftarrow$ use solutions from $\mathcal{A}$ as the true training data and labels. \\
    $\hat{X}$ $\leftarrow$ one-hot encodes the training data $\hat{X}$.\\
    \textcolor{gray}{// Prepare a set of surrogate models for synthetic data generation.} \\
    \{$rbf$, $dt$, $gb$\} $\leftarrow$ fit each model from $\hat{X}$ and $\hat{Y}$.\\
    $\tilde{f}$ $\leftarrow$ initialize a neural network with random weights $w$. \\
    $t$ $\leftarrow$ 0 \textcolor{gray}{// initialize an epoch counter}.\\
    \While{$t < T$}{
        $\eta$ $\leftarrow$ $\frac{1}{2}\eta_{o}\Big(1 + \cos{\left(\frac{t}{T}\pi\right)}\Big)$ \textcolor{gray}{// anneal the learning rate}.\\
        \textcolor{gray}{// Enrich training set by data with synthetic labels.} \\
        $\tilde{X}$ $\leftarrow$ randomly sample a small set of data.\\
        $\tilde{Y}$ $\leftarrow$ use predictions from \{$rbf$, $dt$, $gb$\} as labels for $\tilde{X}$.\\
        $X$ $\leftarrow$ $\hat{X}$ $\cup$ $\tilde{X}$; $Y$ $\leftarrow$ $\hat{Y}$ $\cup$ $\tilde{Y}$\\
        \textcolor{gray}{// Train $\tilde{f}(w)$ with the proposed ranking loss.} \\
        $\mathcal{L}$ $\leftarrow$ calculate the loss on ($X$, $Y$) following Eq~(\ref{def:loss}).\\
        $\nabla w$ $\leftarrow$ Compute the gradient by $\partial \mathcal{L} / \partial w$. \\
        $w$ $\leftarrow$ $(1 - \lambda)w - \eta\nabla w$ \textcolor{gray}{// stochastic gradient descent}.\\
        $t$ $\leftarrow$ $t + 1$\\
    }
\textbf{Return} Trained accuracy predictor $\tilde{f}$($w$). 
\caption{\small Segmentation accuracy predictor (RankNet)\label{algo:ranknet}}
\end{algorithm}

\subsection{Surrogate Modeling for Latency\label{sec:lut}}
To design architectures with real-time performance, we need to include the run-time latency as an additional objective to be minimized. 
However, getting a stable measurement of the latency is not straightforward, as it critically depends on the operating conditions of the hardware, e.g., current loading, ambient temperature, etc. 
To circumvent the issue of inconsistent latency readings and avoid replicating the exact computing environment for every node that one may wish to parallelize the evaluations on, we adopt the look-up table approach~\cite{cai2018proxylessnas}. 

Recall that the architectures encoded in our search space follow feedforward sequential connections at the layer level, i.e., no connections skipping over intermediate layers or an input feeding into multiple layers in parallel. 
The expected latency of an architecture can be approximated as:

\begin{equation}
\small
\begin{aligned}
\mathbb{E}\big[\mbox{latency}\big] = \sum_{l \in L}\underbrace{lat(m_l, e_l) + lat_{s}(d_0, m_0)}_{\substack{\mbox{encoder, stem} \\ \mbox{(searched)}}} + \underbrace{lat_{t} + lat_{d}}_{\substack{\mbox{tail, decoder} \\ \mbox{(common to all)}}},
\end{aligned}
\label{def:latency}
\normalsize
\end{equation}

\noindent where $L$ is the set of active layers specified by the \emph{depth} encoding; $lat(m_l, e_l)$ is the latency of the $l$th layer of the searched encoder network, depending on the output size $m_l$ and the hidden size $e_l$. 
Since the tail, and decoder networks are commonly shared among all searchable architectures, the latency induced by these components act as constant factors to the latency calculation. 

\begin{figure}[!hbt]
    \centering
    \begin{subfigure}[b]{0.24\textwidth}
    \centering
    \includegraphics[width=\textwidth{}]{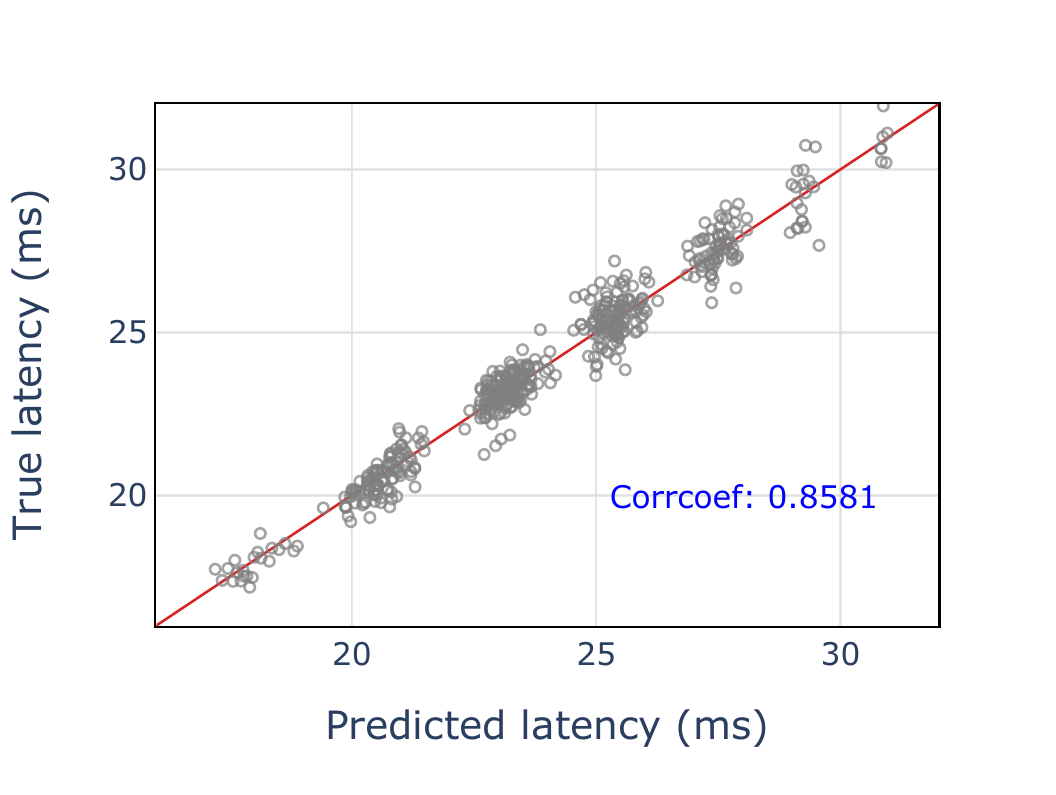}
    \end{subfigure} \hfill
    \centering
    \begin{subfigure}[b]{0.24\textwidth}
    \centering
    \includegraphics[width=\textwidth{}]{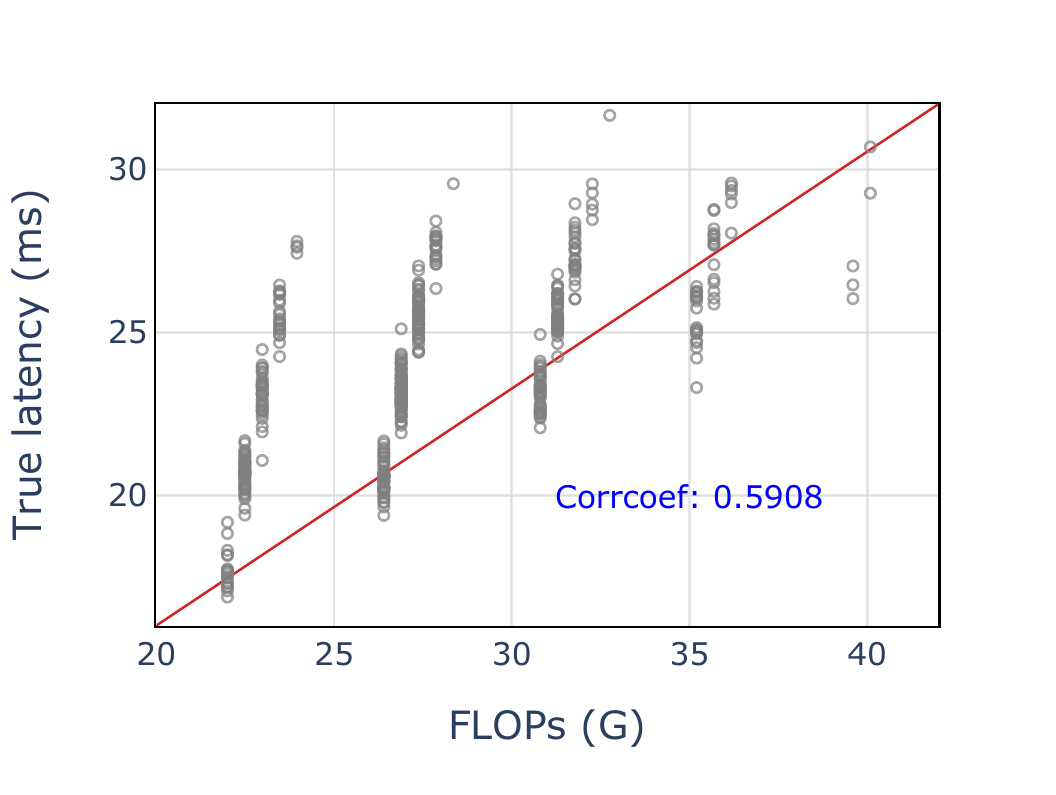}
    \end{subfigure}
\caption{\textbf{Left: Performance of our latency look-up table}. The Kendall rank correlation coefficient $\tau$ is provided along with the fitted linear model (\textcolor{red}{red} line). Correlation from the number of floating-point operations, FLOPs (G), is also provided for comparison (\emph{Right}). \label{fig:lut}}
\end{figure}

The latency of each layer is measured by enumerating over all combinations of $m_l$ and $e_l$ settings (nine options in total assuming independence among layers). 
Then the measured latency is stored in a \emph{look-up table}, through which the latency of a layer under any setting can be queried via a designated key. 
This approach essentially decouples hardware inference from latency computation, allowing the search of optimal architectures to be distributed across multiple nodes without sophisticated control of compute environment. 
A performance overview of the constructed look-up table is provided in Fig.~\ref{fig:lut}.

\subsection{Pre-screening\label{sec:pre-screen}}
With the constructed surrogate accuracy predictor RankNet (Section~\ref{sec:acc_surrogate}) and the latency look-up table (Section~\ref{sec:lut}), the problem presented in Eq~(\ref{def:nas}) essentially becomes a numerical multi-objective optimization problem with discrete variables, which can be exhaustively optimized by conventional multi-objective evolutionary algorithms (MOEAs) in minutes. 
In this work, we adopt NSGA-II~\cite{nsga2} to simultaneously optimize the predicted segmentation accuracy and the predicted latency (line 11 in Algo~\ref{algo:framework}). At the conclusion of the NSGA-II search, a set of non-dominated architectures is returned. 
With the search efficiency in mind, a small subset of the returned candidate arechitectures can be evaluated (with high-fidelity). 
The procedure to optimally select the subset is widely studied in the surrogate modeling literature. 
However, most existing pre-screening (also referred as infill criteria) techniques are  assuming equal evaluation-cost of both objectives. 
While in our case, (high-fidelity) evaluation of one of the objectives (i.e. latency) is significantly cheaper than the other (i.e., segmentation accuracy) in terms of simulation time. 
Thereby, a hierarchical pre-screening procedure is proposed with an inductive bias towards the latency objective. 

As shown in Algorithm~\ref{algo:pre-screen}, the proposed pre-screening routine first evaluates the latency of all the candidate solutions. 
Then the solutions are selected solely based on the latency, to encourage uniformity along the latency objective axis. 
To quantitatively measure the uniformity, we use the Kolmogorov–Smirnov statistic (denoted as $f_{k-s}$), which computes a distance between the empirical cumulative distribution of samples and a reference distribution (i.e., a uniform distribution in our case). 
We formulate this problem as a subset selection problem, as follows:

\begin{algorithm}[t]
\SetAlgoLined
\SetKwInOut{Input}{Input}
\SetKwInOut{Output}{Output}
\SetKwFor{For}{for}{do}{end for}
\footnotesize
\Input{Candidate solutions $Q$, parent population $P$, \# of solutions to be selected $K$.}
    \textcolor{gray}{// First select solutions based on (high-fidelity evaluated) latency.} \\
    $P^*$ $\leftarrow$ get non-dominated (ND) solutions from $P$ via ND sorting.\\
    $Lat_{P^*}$ $\leftarrow$ get latency of ND parent solutions. \\
    $Lat_{Q^{}}$ $\leftarrow$ (high-fidelity) evaluate the latency of candidate solutions.\\
    $Q^*$ $\leftarrow$ select a subset from $Q$ by solving the problem in Eq~\ref{def:subset}.\\
    \textcolor{gray}{// Then fill the remaining slots based on (predicted) accuracy.} \\
    \If{$|Q^*| < K$}{
        $\tilde{Q}$ $\leftarrow$ sort $Q$ in descending order by predicted accuracy. \\
        \For{$q$ in $\tilde{Q}$}{
            \lIf{$q$ $\notin$ $Q^*$}{$Q^*$ $\leftarrow$ $Q^* \cup q$}
            \lIf{$|Q^*| \geq K$}{break}
        }
    }
\textbf{Return} A selected subset for high-fidelity evaluations $Q^*$. 
\caption{Hierarchical pre-screening procedure\label{algo:pre-screen}}
\end{algorithm}

\begin{equation}
\small
\begin{aligned}
\Minimize_{\bm{x}} & \hspace{3mm} f_{k\mbox{-}s}\big(F(L \cdot \bm{x}), F_{\mathcal{U}}\big) , \\
\st  & \hspace{3mm} |\bm{x}| \leq K; \hspace{3mm} x_i \in \{0, 1\}.
\end{aligned}
\label{def:subset}
\normalsize
\end{equation}
\noindent where $L$ represents the list of candidate solutions; 
$\bm{x}$ are binary decision variables where one indicates a solution is selected; 
$F(L \cdot \bm{x})$ is the empirical cumulative distribution function estimated from the selected solutions;
$F_{\mathcal{U}}$ is the cumulative distribution function of the uniform distribution; $K$ is a hyperparameter indicating the maximum \# of solutions to be selected. 

To efficiently solve this subset selection problem, we adopt a binary genetic algorithm with customized crossover and mutation operators to ensure that feasibility is always satisfied during solution generation steps\footnote{More details and pseudocodes are provided in the supplementary materials under Section~1.}. 
In case the optimized number of selected candidate solutions is smaller than the pre-specification of $K$, the rest of the slots are filled according to the predicted segmentation accuracy.

\section{Experiments\label{sec:exper}}
In this section, we first introduce the benchmark datasets and baselines studied in this work, followed by the implementation details. 
We then present the empirical results to evaluate the efficacy of \ourmethod{}. 
Further analysis on the transferability of the obtained models is also studied. 

\subsection{Datasets}
We consider three widely-studied benchmark datasets to evaluate our method, including Cityscapes \cite{cityscapes}, COCO-Stuff-10$K$ \cite{caesar2018coco}, and PASCAL VOC 2012 \cite{pascal-voc}. These three datasets cover a diverse variety of images in terms of objects, scenes, and scales for semantic segmentation. Examples from these datasets are provided in Fig.~\ref{fig:city-visualization} and Fig.~\ref{fig:coco-visualization} for visualization. 

Cityscapes \cite{cityscapes} is a large-scale dataset for semantic understanding of urban street scenes. It is officially split into a training set of 2,975 images, a validation set of 500 images, and a (privately hosted) testing set of 1,525 images. The provided annotations (i.e., ground truth labels) include 30 classes, 19 of which are used for semantic segmentation. The images in Cityscapes are of higher (and unified) spatial resolution of $1024\times2048$ pixels, which make the dataset a challenging task for real-time semantic segmentation. In this paper, we only use images with fine annotations\footnote{There are roughly 20,000 additional coarsely annotated training images available.} to train and validate our proposed method. 

COCO-Stuff-10$K$ \cite{caesar2018coco} is a subset of 10$K$ images from the original MS COCO dataset \cite{mscoco} with additional dense stuff annotations. COCO-Stuff-10$K$ is also a challenging task for semantic segmentation as it contains 182 different semantic categories---91 categories each for thing and stuff classes, respectively. In this paper, we follow the official split of 9,000 images for training and 1,000 images for testing. 

PASCAL VOC 2012 \cite{pascal-voc} is a relatively small-scale dataset containing 20 foreground object categories and one background class. To be consistent with prior works \cite{liu2019auto}, we augment the original training set with the extra annotations provided by \cite{6126343}, resulting in 10,582 images (\emph{train\_aug}) in total for training.

\subsection{Baselines}
Representative semantic segmentation methods and models are selected from the literature to evaluate the effectiveness of the proposed method. The chosen peer competitors can be broadly classified into two categories depending on if the architectures are manually designed by skilled practitioners or automatically generated by algorithms. The first group of human engineered models includes BiSeNet \cite{yu2018bisenet}, ESPNetv2 \cite{mehta2019espnetv2}, and HRNet \cite{hrnetv2}, etc. The second category consists of models that are optimized both by reinforcement learning (RL) based methods (e.g., AuxCell \cite{nekrasov2019fast}) and continuous relaxation (gradient) based methods (e.g., Auto-DeepLab \cite{liu2019auto}, CAS \cite{zhang2019customizable}). 

For real-time semantic segmentation, the effectiveness of each model is evaluated by both the segmentation accuracy and the inference speed (i.e., reciprocal of latency). We adopt the metric of mean Intersection-over-Union (mIoU) to evaluate the segmentation accuracy. It computes the IoU for each semantic class, which is then averaged over classes. And the IoU metric measures the number of pixels common between the ground truth and predicted masks divided by the total number of pixels present across both masks, mathematically as below:
$$\mbox{mIoU} = \frac{1}{K+1}\sum_{i=0}^{K}\frac{p_{ii}}{\sum_{j=0}^{K}p_{ij} + \sum_{j=0}^{K}p_{ji} - p_{ii}} $$
where $K$ denotes the total number of foreground classes; $p_{ij}$ counts the number of pixels whose ground-truth labels are $i$, but are predicted as $j$. 

\subsection{Implementation Details\label{sec:implement}}
We pretrain the hypernetwork on ImageNet-1K \cite{imagenet} for 1,200 epochs, followed by 300 epochs of fine-tuning on Cityscapes \cite{cityscapes}. To stabilize the hypernetwork training, the searched architectural options (e.g., \# of layers, \# of channels, etc.) are gradually activated during the training process \cite{onceforall}. This procedure takes about a week on a server with eight Titan RTX GPUs. It's worth mentioning that the majority of the computation expense is from ImageNet-1K pretraining, which is a one-time cost that can be amortized over many search runs. For practicality consideration, the search is carried out on Cityscapes. A subset of 500 images are randomly separated from the training set to guide the evolution. We run \ourmethod{} with two objectives: maximize the mIoU and inference speed. This search itself takes less than a day on a single Titan RTX GPU card, and is repeated five times to capture the variance in performance induced by the stochastic nature of the algorithm. We select and report the performance of the median run as measured by the hypervolume (HV) calculated from the reference point being at the origin. Table~\ref{tab:hyperparameter} summarizes the hyperparameter settings that were guided by the ablation experiments described in Section~\ref{sec:ablation}. 

\begin{table}[t]
\centering
\caption{Summary of hyperparameter settings.\label{tab:hyperparameter}}
\resizebox{.4\textwidth}{!}{%
\begin{tabular}{@{\hspace{2mm}}l|l|c@{\hspace{2mm}}}
\toprule
Category & Parameter & Setting \\ \midrule
\multirow{3}{*}{Global} & population size & 300 \\
 & \# of generations & 20 \\
 & \# of high-fidelity eval. per gen. & 8 \\ \midrule
\multirow{4}{*}{\begin{tabular}[c]{@{}l@{}}Accuracy predictor\\(RankNet)\end{tabular}} & \# of training epochs & 500 \\
 & \# of layers / neurons & 3 / 300 \\
 & weight decay & $\num{1e-5}$ \\
 & init. learning rate & $\num{8e-4}$ \\ 
 & margin $\gamma$ & $0.05$ \\ 
 \midrule
\multirow{4}{*}{\begin{tabular}[c]{@{}l@{}}NSGA-II\end{tabular}} & population size & 100 \\
 & \# of generations & 1,000 \\
 & crossover probability & 0.9 \\
 & mutation probability & 0.05 \\ 
 \bottomrule
\end{tabular}%
}
\end{table}

\subsection{Results on Cityscapes}
\begin{figure}[t]
    \centering
    \begin{subfigure}{0.48\textwidth}
    \centering
    \includegraphics[width=0.9\textwidth{}]{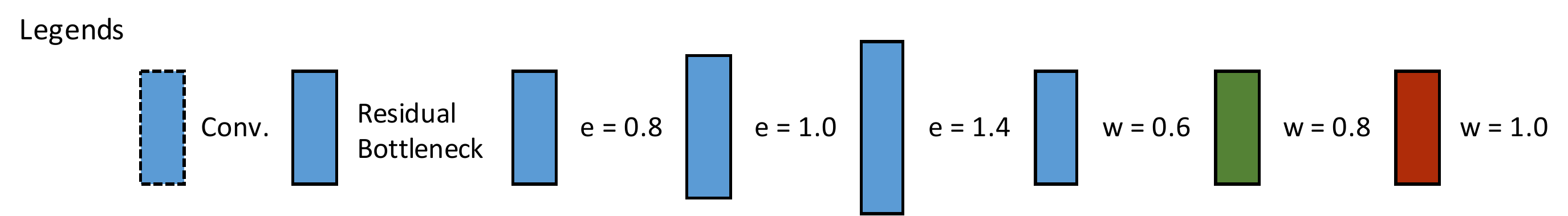} \\
    \includegraphics[width=0.98\textwidth{}]{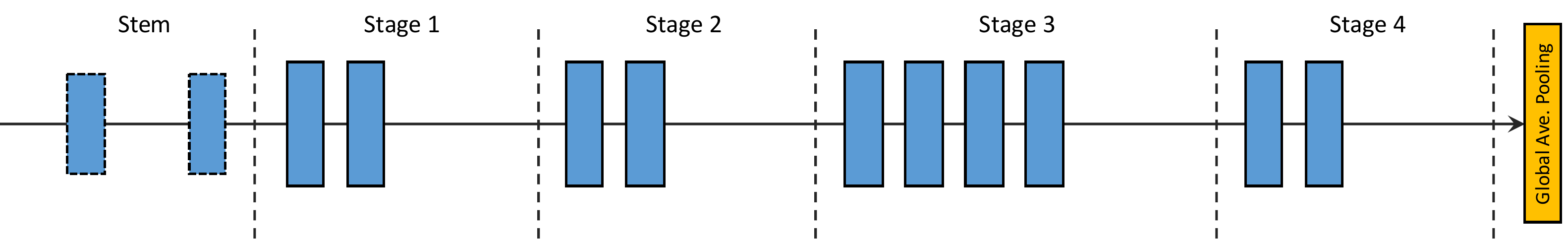} \\
    \includegraphics[width=0.98\textwidth{}]{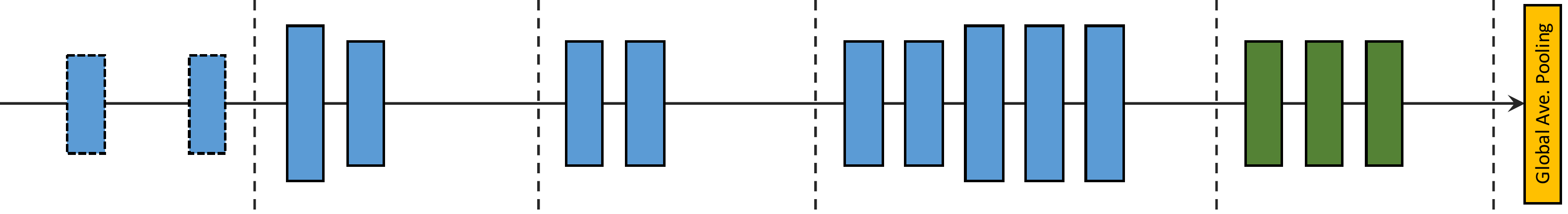} \\
    \includegraphics[width=0.98\textwidth{}]{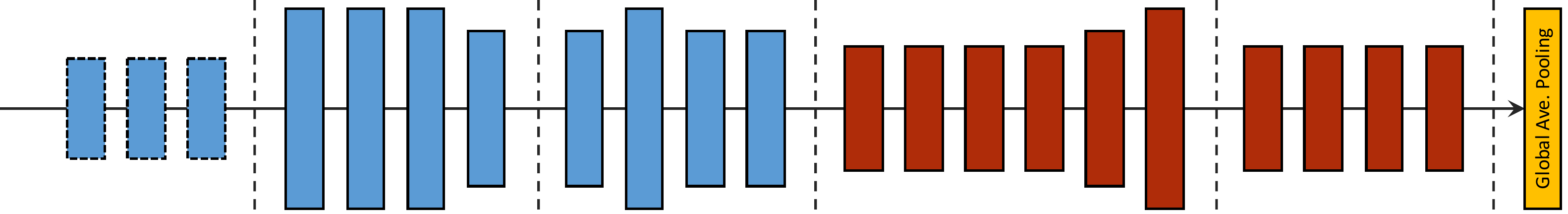}
    \end{subfigure}
\caption{MoSegNets-\{small, median, large\} Architectures from Trade-Off Front. Architectures are arranged in descending inference-speed order from top to bottom. \label{fig:mosegnets}}
\vspace{-0.3cm}
\end{figure}

From the set of non-dominated solutions returned at the conclusion of the evolution, we choose five architectures |based on their trade-offs (following the procedures outlined in supplementary materials under Section~1). For reference, our obtained models are referred as \ourmodel{}s (Fig.~\ref{fig:mosegnets}). For comparison with other peer methods, we re-train the weights of each model thoroughly from scratch. Our post-search training largely follows \cite{yu2018bisenet}: SGD optimizer with momentum 0.9 and weight decay $\num{1e-5}$; batch normalization and size of eight images per GPU card. The learning rate decays following a ``poly'' schedule (i.e., $0.01 \times (1 - \frac{iter}{maxIter})^{0.9}$) from 0.01 to zero in 40$K$ iterations\footnote{Each iteration refers to each mini-batch.}. The data augmentation settings include horizontal flip, scale, and crop to $1536 \times 768$. The scaling ratio is randomly selected from \{0.75, 1.0, 1.25, 1.5, 1.75, 2.0\}. We note that our reported inference speed is measured on a single Titan RTX GPU card without optimized inference acceleration implementations (e.g., TensorRT\footnote{\url{https://developer.nvidia.com/tensorrt}}); and our reported mIoU is computed without test-time augmentation techniques (e.g., multi-crop, multi-scale) unless otherwise specified. 

\begin{table}[t]
\centering
\caption{\textbf{Comparison with state-of-the-art models on Cityscapes} \cite{cityscapes}. Models are grouped into sections and the best result in each section is in bold. ``RL'' stands for reinforcement learning.  $^*$ denotes the use of high-performance inference framework \emph{TensorRT} to measure speed. $^\ddagger$ is an ensemble of multiple input scales. \label{tab:city-main}}
\resizebox{.46\textwidth}{!}{%
    \begin{tabular}{@{\hspace{2mm}}l@{\hspace{-1mm}}l|c|cc|c@{\hspace{2mm}}}
    \toprule
      & Model & Method & \begin{tabular}[c]{@{}c@{}} FLOPs $\downarrow$\\ (G)\end{tabular} & \begin{tabular}[c]{@{}c@{}} Speed $\uparrow$\\ (imgs / sec)\end{tabular} & \begin{tabular}[c]{@{}c@{}}mIoU $\uparrow$\\ (\%)\end{tabular} \\ \midrule
     \parbox[t]{6mm}{\multirow{9}{*}{\rotatebox[origin=c]{90}{\emph{real-time models}}}} & ESPNetV2 \cite{mehta2019espnetv2} & \multirow{4}{*}{manual} & - & - & 66.4 \\
     & BiSeNet \cite{yu2018bisenet} &  & 72 & 65.5 & 74.8 \\
     & SwiftNet \cite{orsic2019defense} &  & 104 & 39.9 & 75.4 \\
     & BiSeNetV2 \cite{yu2020bisenet} &  & 78 & 47.3 & 75.8 \\
     & CAS \cite{zhang2019customizable} & \multirow{2}{*}{gradient} & - & \textbf{108}$^*$ & 71.6 \\
     & FNA \cite{Fang2020Fast} &  & \textbf{24} & 40.8 & 76.6 \\
     & AuxCell \cite{nekrasov2019fast} & RL & - & 44.2 & 77.1 \\\cmidrule{2-6}
     & \textbf{\ourmodel{}}$_{small}$ & \multirow{2}{*}{EA} & 35 & 73.2 & 75.9 \\
     & \textbf{\ourmodel{}}$_{large}$  &  & 42 & 50.1 & \textbf{78.2} \\ \midrule\midrule
     \parbox[t]{4mm}{\multirow{6}{*}{\rotatebox[origin=c]{90}{\emph{large models}}}} & PSPNet \cite{zhao2017pyramid} & \multirow{3}{*}{manual} & 412 & 0.8 & 78.4 \\
     & DeepLabv3+ \cite{chen2018encoder} &  & - & - & 79.6 \\
     & HRNetV2 \cite{hrnetv2} &  & 696 & \textcolor{black}{6.3} & 81.1 \\
     & Auto-DeepLab-S \cite{liu2019auto} & gradient & 333 & - & 79.7 \\\cmidrule{2-6}
     & \textbf{\ourmodel{}}$_{xlarge}$ & \multirow{2}{*}{EA} & \textbf{128} & \textbf{29.2} & 79.5 \\
     & \textbf{\ourmodel{}}$_{xlarge}^\ddagger$ &  & - & 0.5 & \textbf{81.3} \\ \bottomrule
    \end{tabular}%
    }
\end{table}

\begin{figure}[t]
    \centering
    \begin{subfigure}{0.48\textwidth}
    \centering
    \includegraphics[width=0.95\textwidth{}]{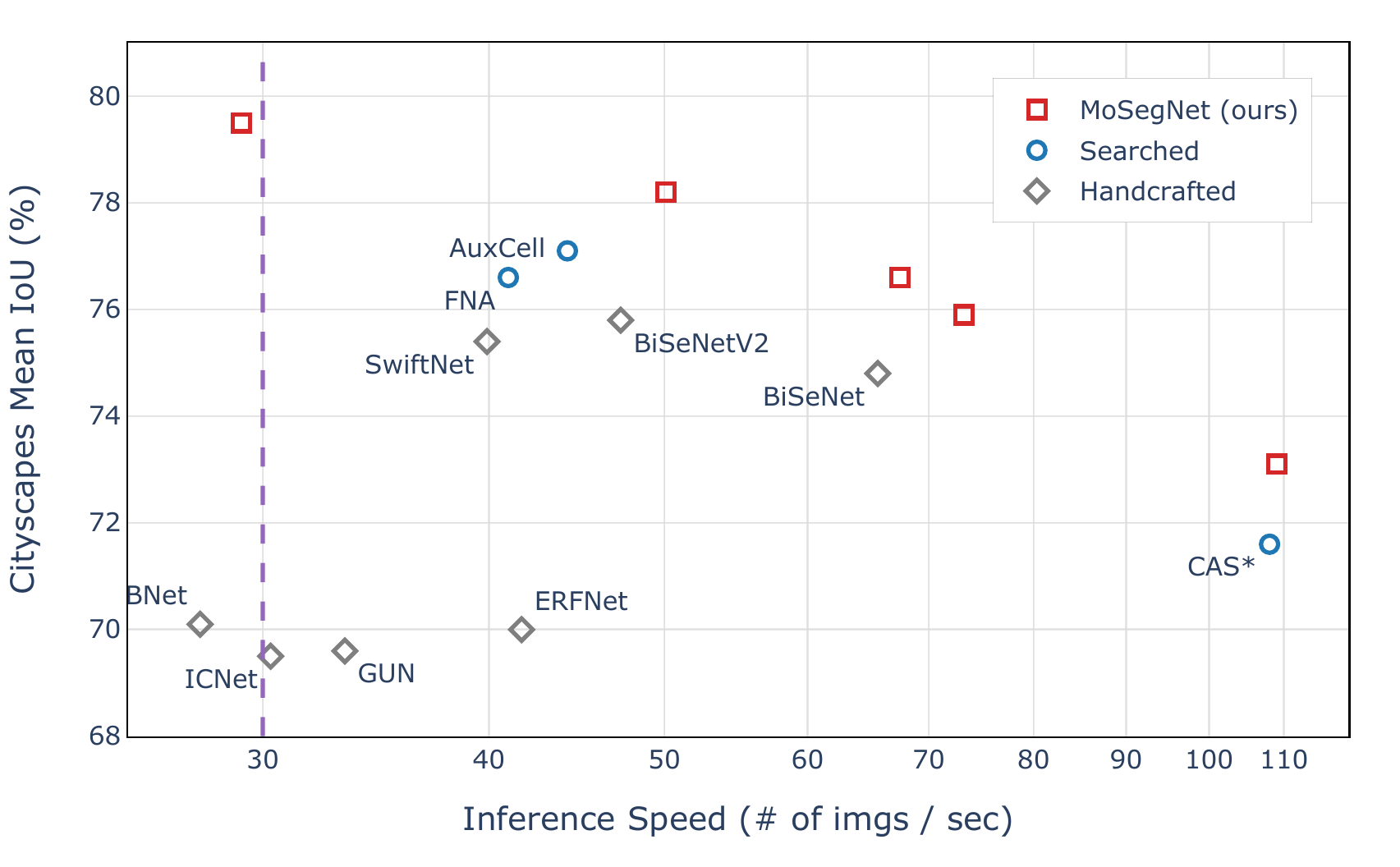}
    \end{subfigure}
\caption{\textbf{\emph{Speed-Accuracy} trade-off comparison} with other real-time models on Cityscapes. Models with inference speed of more than 30 images per second (\textcolor{violet}{purple} line) are considered to be real-time. $^*$ uses high-performance inference framework \emph{TensorRT} to measure speed. See Table~\ref{tab:city-main} for more comparisons to large models (non-real-time).\label{fig:city-realtime}}
\end{figure}

\begin{figure*}[!hbt]
    \centering
    \begin{subfigure}{0.95\textwidth}
    \centering
    \includegraphics[width=0.95\textwidth{}]{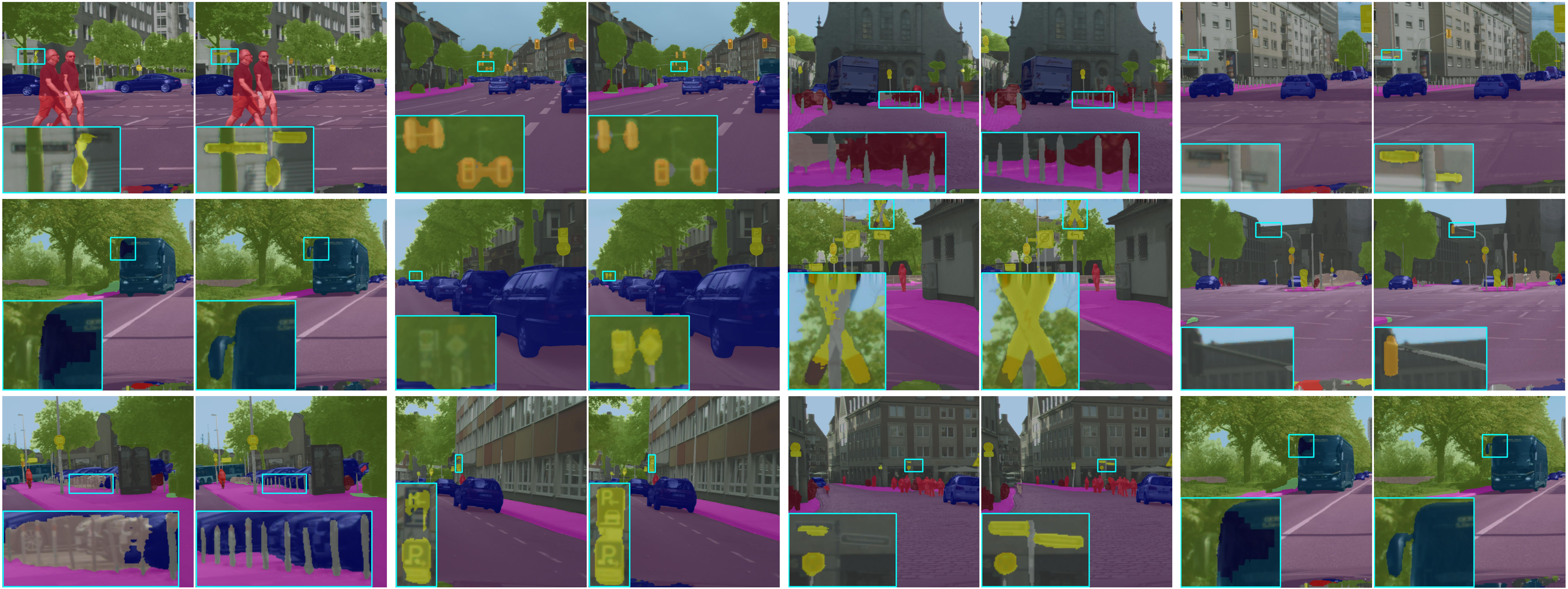}
    \end{subfigure}
\caption{\textbf{Qualitative comparison on Cityscapes} \cite{cityscapes}. Example result pairs visualizing the segmentation performance of BiSeNet \cite{yu2018bisenet} (\emph{left} image) vs. our proposed \ourmodel{} (\emph{right} image). Note how \ourmodel{} predicts masks with substantially finer detail on small objects. \label{fig:city-visualization}}
\end{figure*}

Table~\ref{tab:city-main} presents the experimental results in terms of model efficiency (FLOPs and inference speed) and segmentation accuracy (mIoU) of the compared models. Specifically, Table~\ref{tab:city-main} is divided into two sections for comparison between real-time and large models. The compared models are further grouped based on methods, and are in ascending mIoU order. The symbol ``-'' is used to denote the information is not available publicly. \revision{In general, results indicate that our models perform favorably against a wide range of models that were designed manually or generated by algorithms. In particular, with the same decoder, \ourmodel{}$_{small}$ improves BiSeNet \cite{yu2018bisenet} in all aspects, i.e., 1.1 points more accurate in mIoU, 10\% faster per second in inference, and 2$\times$ more efficient in FLOPs. While \ourmodel$_{large}$ is on average 1.3 points more accurate and 18\% faster than models obtained by RL-based methods, i.e., FNA \cite{Fang2020Fast} and AuxCell \cite{nekrasov2019fast}. When compared to non real-time models, \ourmodel{}$_{xlarge}$ is 36$\times$ faster than PSPNet \cite{zhao2017pyramid} while being more accurate in mIoU;  it is also 3$\times$ more efficient in FLOPs than Auto-DeepLab-S \cite{liu2019auto} while being comparable in mIoU; and it achieves state-of-the-art performance in Cityscapes mIoU with multi-scale inference, surpassing HRNetV2 \cite{hrnetv2}. Moreover, Figs.~\ref{fig:city-realtime} and \ref{fig:city-visualization} provide visual comparisons on speed-accuracy trade-off curve and sample outputs, respectively. 
}

\subsection{Transferability to other Datasets}
In line with the practice adopted in most previous NAS methods \cite{liu2019auto,zhang2019customizable,nekrasov2019fast,Fang2020Fast}, we evaluate the transferability of the obtained architectures by inheriting the topology optimized for one dataset with weights retrained for a new dataset. In this work, we apply the evolved \ourmodel{}s on Cityscapes dataset \cite{cityscapes} to two other datasets---i.e., COCO-Stuff-10$K$ \cite{caesar2018coco} and PASCAL VOC 2012 \cite{pascal-voc}. And the retraining procedure for these two datasets is very similar to the one used for Cityscapes. The cropped image size is modified to 640$\times$640 for COCO-Stuff and 512$\times$512 for PASCAL VOC. The number of training iterations is halved for COCO-stuff and increased by 20$K$ for PASCAL VOC. 

\begin{table}[!hbt]
\centering
\caption{\textbf{Performance comparison on COCO-Stuff-10$\bm{K}$} \cite{caesar2018coco}. Models are evaluated on network complexity (FLOPs), inference speed, and accuracy (mIoU). The best result in each section is in bold. \label{tab:coco-main}}
\resizebox{.45\textwidth}{!}{%
    \begin{tabular}{@{\hspace{2mm}}l|cc|c@{\hspace{2mm}}}
    \toprule
    Model & FLOPs $\downarrow$ (G) & Speed $\uparrow$ (imgs / sec) & mIoU $\uparrow$ (\%) \\ \midrule
    FCN \cite{long2015fully} & - & 5.9 & 22.7 \\
    DeepLab & - & 8.1 & 26.9 \\
    BiSeNet \cite{yu2018bisenet} & 25.1 & - & 28.1 \\
    BiSeNetV2 \cite{yu2020bisenet} & 27.1 & 42.5 & 28.7 \\
    ICNet \cite{zhao2018icnet} & - & 35.7 & 29.1 \\
    PSPNet50 \cite{zhao2017pyramid} & - & 6.6 & 32.6 \\ \midrule
    \textbf{\ourmodel{}}$_{small}$ & \textbf{12.0} & \textbf{92.1} & 30.5 \\
    \textbf{\ourmodel{}}$_{large}$ & 14.3 & 78.3 & \textbf{34.2} \\ \bottomrule
    \end{tabular}%
}
\end{table}

\begin{table}[!hbt]
\centering
\caption{\textbf{Performance comparison on PASCAL VOC 2012} \cite{pascal-voc}. Models are evaluated on network complexity (FLOPs) and accuracy (mIoU). ``COCO'' indicates additional pretraining on MS COCO dataset \cite{mscoco}. The best result in each section is in bold. \label{tab:voc-main}}
\resizebox{.47\textwidth}{!}{%
    \begin{tabular}{@{\hspace{2mm}}l|c|c|cc@{\hspace{2mm}}}
    \toprule
    Model & Method & COCO & FLOPs $\downarrow$ (G) & mIoU $\uparrow$ (\%) \\ \midrule
    BiSeNet \cite{yu2018bisenet} & \multirow{5}{*}{manual} &  & 16 & 71.0 \\
    DeepLabv3+ \cite{chen2018encoder} &  &  & 101 & 79.0 \\
    Res2Net \cite{res2net} &  &  & 101 & 80.2 \\
    PSPNet \cite{zhao2017pyramid} &  &  & - & 82.4 \\
    RefineNet \cite{refinenet} &  & \checkmark & 261 & \textbf{82.9} \\
    Auto-DeepLab-S \cite{liu2019auto} & \multirow{2}{*}{gradient} &  & 43 & 71.7 \\
    Auto-DeepLab-L \cite{liu2019auto} &  & \checkmark & 87 & 80.8 \\
    \midrule
    \textbf{\ourmodel{}}$_{small}$ & \multirow{2}{*}{EA} &  & \textbf{7.8} & 75.8 \\
    \textbf{\ourmodel{}}$_{large}$ & & \checkmark & 9.3 & {82.2} \\ \bottomrule
    \end{tabular}%
}
\end{table}

\begin{figure}[!hbt]
    \centering
    \begin{subfigure}{0.45\textwidth}
    \centering
    \includegraphics[width=0.9\textwidth{}]{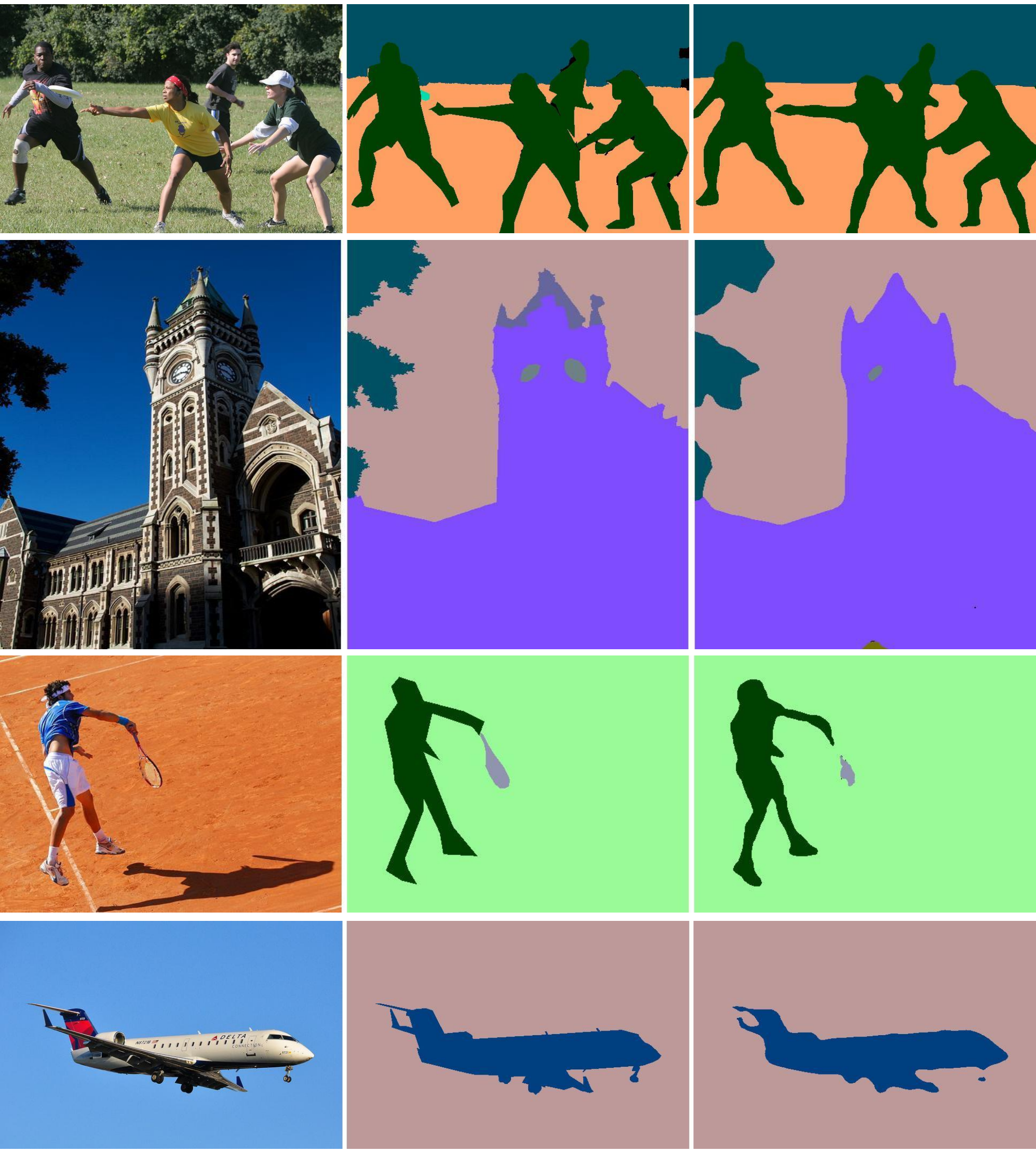}
    \end{subfigure}
\caption{\textbf{Visual Comparison on COCO-Stuff} \cite{caesar2018coco}. Example pairs visualizing the input image (\emph{left}), the ground truth (\emph{middle}), and the predictions from the proposed \ourmodel{} (\emph{right}). \label{fig:coco-visualization}}
\end{figure}

Table~\ref{tab:coco-main} and \ref{tab:voc-main} compare the performance in terms of model complexity (inference speed or FLOPs) and segmentation accuracy (mIoU) on COCO-Stuff and PASCAL VOC, respectively. In general, our models are consistently more accurate than peer competitors while being an order of magnitude more efficient in FLOPs or inference speed, further validating the effectiveness of \ourmethod{} under transfer learning setup. More specifically, \ourmodel{}$_{small}$ is 1.8 - 4.8 points more accurate in mIoU than BiSeNet \cite{yu2018bisenet} on COCO-stuff and PASCAL VOC, while being 2$\times$ more efficient in FLOPs and inference speed. Without pretraining on extra MS COCO data \cite{mscoco}, \ourmodel{}$_{large}$ achieves 2.2 points higher mIoU on PASCAL VOC than peer NAS method Auto-DeepLab-L \cite{liu2019auto} while using an order of magnitude less FLOPs. Example outputs are provided in Fig.~\ref{fig:coco-visualization} for visualization.

\section{Ablation Studies\label{sec:ablation}}
This section aims to disentangle the individual contribution of each main component in the proposed method, followed by the hyperparameter analysis. 

\subsection{Analysis of Surrogate Modeling}

To assess the effectiveness of the adopted surrogate modeling pipeline, we collect a number of well-established surrogate models from the literature. 
The considered surrogate models range from low-complexity methods (e.g., RBF, SVR), to more sophisticated ensemble-based methods (e.g., Gradient Boosting \cite{ke2017lightgbm}), and dedicated DNN performance predictors proposed by previous NAS methods (e.g., E2EPP \cite{e2epp}). 
We uniformly sample from the search space to generate a pool of 1,100 architectures, where 1000 and 100 architectures are selected as the the training set and test set respectively. 
For architectures in the testing set, we train them with SGD optimizer on Cityscapes for 40$K$ iterations and use the segmentation accuracy (mIoU) computed at the end of the training as the ground truth targets. 
For architectures in the training set, we use segmentation accuracy computed with the inherited weights from the hypernetwork as the training targets. 

Recall that the selection of architectures in our proposed evolutionary routine compares relative performance differences among architectures. 
Hence, we adopt coefficients of correlation to quantitatively compare different surrogate models, as oppose to root mean square error or coefficient of determination. 
Specifically, we report all three indicators for measurement of correlation, including Pearson coefficient ($r$), Spearman's Rho ($\rho$), and Kendall's Tau ($\tau$). 
The values of these three indicators range between [-1, 1] with a higher value indicating a better prediction performance. 
Table~\ref{tab:surr_model} presents the experimental results comparing our surrogate model with the selected peer competitors. 
Evidently, our method significantly outperforms other methods, achieving the best correlation coefficient in Pearson $r$, Spearman $\rho$ and Kendall $\tau$. 

\begin{table}[!hbt]
\centering
\caption{\textbf{Coefficients of correlation comparison}. Each method is trained with 1,000 samples and evaluated on a held-out test set of 100 samples. The results are averaged over 31 runs with standard deviation shown in the parentheses. ``Time'' denotes the training time measured in seconds. The best result in each section is in bold. \label{tab:surr_model}}
\resizebox{.47\textwidth}{!}{%
\begin{tabular}{@{\hspace{2mm}}l|c|c|c|c@{\hspace{2mm}}}
\toprule
Method & Pearson $r$ & Spearman $\rho$ & Kendall $\tau$ & Time (sec.) \\ \midrule
RBF & 0.6199 (0.12) & 0.5688 (0.10) & 0.4692 (0.05) & 0.02 \\
Kriging & 0.1161 (0.18) & 0.0211 (0.15) & 0.0107 (0.10) & 4.52 \\
SVR & 0.5721 (0.11) & 0.5597 (0.10) & 0.4386 (0.05) & 0.11 \\
DT & 0.6621 (0.15) & 0.5986 (0.10) & 0.5377 (0.05) & \textbf{0.01} \\
GB \cite{ke2017lightgbm} & 0.7409 (0.10) & 0.7079 (0.08) & 0.6259 (0.04) & 0.61 \\
E2EPP \cite{e2epp} & 0.7746 (0.10) & 0.7302 (0.07) & 0.6162 (0.04) & 1.01 \\ \midrule
\textbf{RankNet (ours)} & \textbf{0.7988 (0.08)} & \textbf{0.7658 (0.07)} & \textbf{0.6873 (0.03)} & 3.82 \\ \bottomrule
\end{tabular}%
}
\end{table}

\begin{figure}[!hbt]
    \centering
    \begin{subfigure}[b]{0.45\textwidth}
    \centering
    \includegraphics[width=.95\textwidth{}]{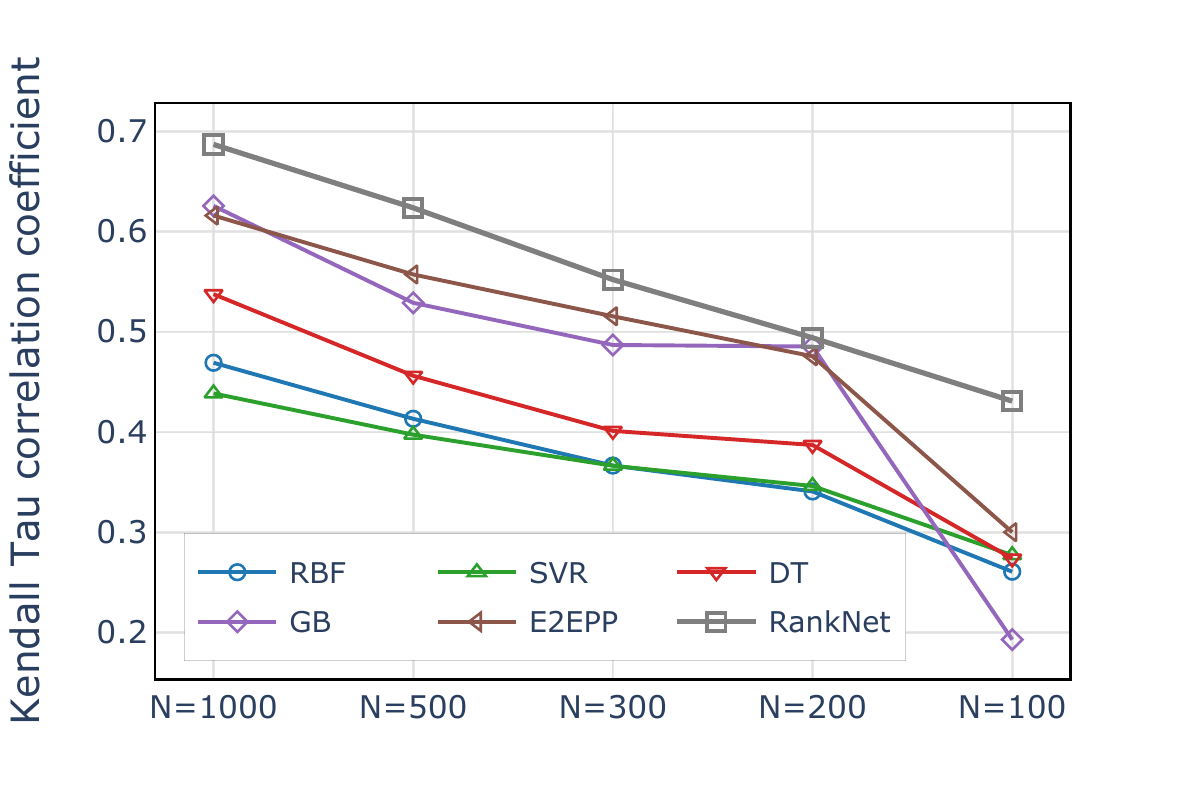}
    \end{subfigure}
\caption{\textbf{Kendall rank-order correlation} comparison under different \# of training samples $N$. For each method, we repeat for 31 runs and report the mean performance.\label{fig:surr_model}}
\end{figure}

In addition to being indicative in prediction, another desired property of a surrogate model is sample efficiency. 
To validate the scalability of our method with respect to the training size, we repeat the previous experiment with gradually reduced \# of training samples. 
The experimental results are presented in Fig.~\ref{fig:surr_model}. 
In general, we observe that all methods exhibit a significant degradation in performance, i.e., lower mean and higher variance in Kendall $\tau$ as the \# of available training samples reduces. 
However, our method remains consistently better than compared methods across all training size regimes, particularly in the high sample-efficient regime. 

Towards facilitating a better understanding on the observed effectiveness and efficiency, we visualize the development progress of our proposed surrogate model in Table~\ref{tab:development}, showing the impacts of the main components introduced in RankNet. We quantitatively measure the impacts by the Kendall $\tau$ correlation coefficient and the time complexity. 
A method with higher $\tau$ and lower time complexity is preferred. 

\begin{table}[!hbt]
\centering
\caption{\textbf{Development of RankNet}. We disentangle the impact of adding individual components to a multi-layer perceptron (MLP; baseline), leading to our proposed surrogate model, i.e. RankNet. ``GPU'' indicates the use of GPU (CUDA) for training acceleration. All configurations are with 100 training samples.\label{tab:development}}
\resizebox{.45\textwidth}{!}{%
\begin{tabular}{@{\hspace{2mm}}l|cccc|cc@{\hspace{2mm}}}
\toprule
Method & \begin{tabular}[c]{@{}c@{}}One-hot\\ encoding\end{tabular} & \begin{tabular}[c]{@{}c@{}}Ranking\\ loss\end{tabular} & \begin{tabular}[c]{@{}c@{}}Synthetic\\ data\end{tabular} & GPU & Kendall $\tau$ $\uparrow$ & Time (sec.) $\downarrow$ \\ \midrule
baseline &  &  &  &  & 0.0036 & 2.4 \\ \midrule
 & \checkmark &  &  &  & 0.3157 & 2.9 \\
 &  & \checkmark &  &  & 0.2170 & 3.8 \\
 &  &  & \checkmark &  & 0.0770 & 5.4 \\
 & \checkmark & \checkmark &  &  & 0.4180 & 6.7 \\
 & \checkmark & \checkmark & \checkmark &  & 0.4309 & 6.9 \\ \midrule
RankNet & \checkmark & \checkmark & \checkmark & \checkmark & 0.4309 & 3.7 \\ \bottomrule
\end{tabular}%
}
\end{table}

\subsection{Analysis of Pre-screening}
Towards quantifying the effectiveness of the proposed hierarchical pre-screening criterion, the following ablative experiments are performed. 
Specifically, we consider two other conventional pre-screening methods along with a variant of the proposed method, as follows:
\begin{enumerate}
    \item \emph{Hierarchical}: the proposed hierarchical pre-screening procedure described in Algorithm~\ref{algo:pre-screen}.
    \item \emph{Survival}: the candidate solutions are selected based on their ranks and crowding distances, computed by applying NSGA-II environmental survival operator to the merged population of parents and all candidate solutions.
    \item \emph{HV contribution}: the candidate solutions are selected based on their individual contributions to the HV. 
    \item \emph{Latency}: the candidate solutions are selected based on the uniformity along the second objective of latency. 
\end{enumerate}

\begin{figure}[!hbt]
    \centering
    \begin{subfigure}[b]{0.4\textwidth}
    \centering
    \includegraphics[width=1\textwidth{}]{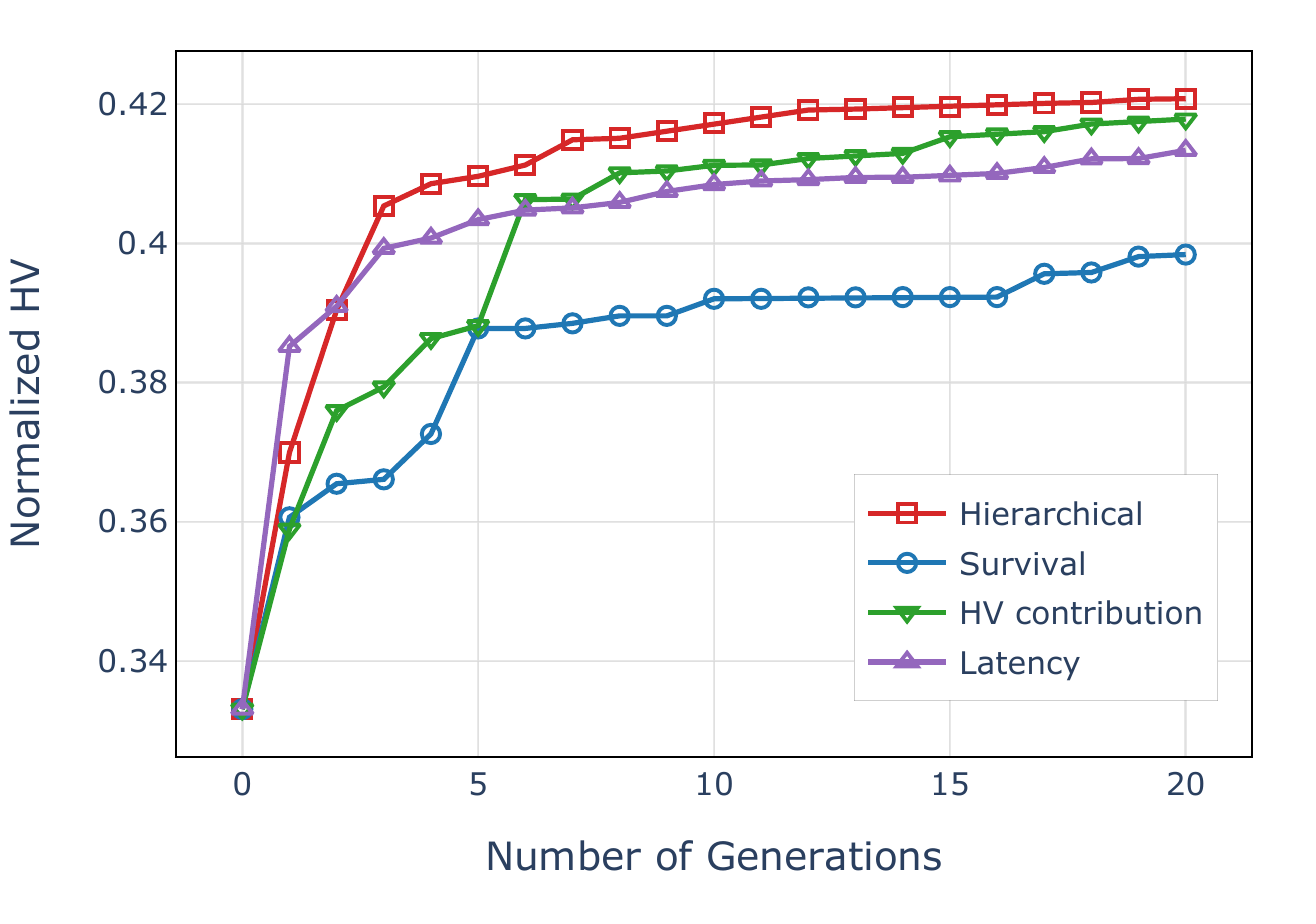}
    \end{subfigure}
\caption{Ablation study on the proposed hierarchical pre-screening criterion. The median generation-wise HV over five runs are reported.\label{fig:pre-screen}}
\end{figure}

On Cityscapes dataset \cite{cityscapes}, we run each setting to maximize both segmentation accuracy and inference speed (reciprocal of latency) for five times and report the median performance (measured in HV) as a function of the number of generations in Fig.~\ref{fig:pre-screen}. 
Evidently, the consistently higher HV suggests that the proposed hierarchical pre-screening criterion is more effective in selecting candidate solutions for high-fidelity evaluation than all other compared methods. 

\subsection{Search vs. Transfer from ImageNet}
Directly using an efficient encoder architecture designed for ImageNet-1K classification (e.g., ResNet \cite{resnet}) has been the defacto method for dense image prediction tasks, including semantic segmentation. 
However, we argue that this method is conceptually sub-optimal given that classification mainly requires architectures to extract higher-level contextual information, while segmentation requires rich spatial details to be simultaneously maintained. 
In line with this derivation, we aim to evaluate the effectiveness of searching directly for segmentation (our method) by comparing the obtained models with efficient models designed for classification by other NAS methods \cite{cai2018proxylessnas,mobilenetv3,wan2020fbnetv2}. 

We collect three state-of-the-art mobile models of relatively the same complexity (in terms of inference speed) as \ourmodel{}$_{small}$, and evaluate all four models for both classification on ImageNet and segmentation performance under the same training hyperparameter settings. 
Specifically, the training on ImageNet follows \cite{cai2018proxylessnas}: RMSProp optimizer with decay 0.9 and momentum 0.9; weight decay $\num{1e-5}$; batch size 512. The learning rate decays from 0.012 to zero following the cosine annealing schedule. 
The data augmentation includes horizontal flip, crop and dropout ratio 0.2. 

The experimental results presented in Table~\ref{tab:development} indicate that the classification and segmentation performance can be in conflict situation, suggesting the necessity of searching directly for the targeted task instead of transferring from ImageNet. 
In particular, despite the deficiency in ImageNet top-1 accuracy, \ourmodel{}$_{small}$ significantly outperforms other three state-of-the-art ImageNet models on all three semantic segmentation datasets. 

\begin{table}[!hbt]
\centering
\caption{Semantic segmentation performance comparison on backbone (encoder) models optimized for classification (\emph{Cls.}) and segmentation (\emph{Seg.}). The compared backbone models are of similar complexity in terms of inference speed. The averaged relative differences are shown in parentheses.\label{tab:transfer-abl}}
\resizebox{.49\textwidth}{!}{%
    \begin{tabular}{@{\hspace{2mm}}l|c|r|rrr@{\hspace{2mm}}}
    \toprule
    \multirow{2}{*}{Backbone} & \multirow{2}{*}{\begin{tabular}[c]{@{}c@{}}Optimized \\ for\end{tabular}} & \multirow{2}{*}{\begin{tabular}[c]{@{}c@{}}ImageNet \\ Top-1 \end{tabular}} & \multicolumn{3}{c}{Semantic Segmentation mIoU (\%)} \\ \cmidrule(l){4-6} 
     &  &  & Cityscapes & VOC 2012 & COCO-Stuff \\ \midrule
    ProxylessNAS \cite{cai2018proxylessnas} & Cls. & 74.6 & 73.4 & 74.2 & 28.2 \\
    MobileNetV3 \cite{mobilenetv3} & Cls. & 75.2 & 75.2 & 73.8 & 28.5 \\
    FBNetV2 \cite{wan2020fbnetv2} & Cls. & 75.5 & 72.6 & 73.6 & 28.5 \\
    \textbf{\ourmodel{}}$_{small}$ & Seg. & (\textcolor{blue}{-1.7}) 73.4 & (\textcolor{red}{+2.9}) 75.9 & (\textcolor{red}{+1.9}) 75.8 & (\textcolor{red}{+2.1}) 30.5 \\ \bottomrule
    \end{tabular}%
}
\end{table}

\section{Application to Huawei Atlas 200 DK\label{sec:application}}

\begin{figure}[t]
    \centering
    \begin{subfigure}[b]{0.45\textwidth}
    \centering
    \includegraphics[width=1\textwidth{}]{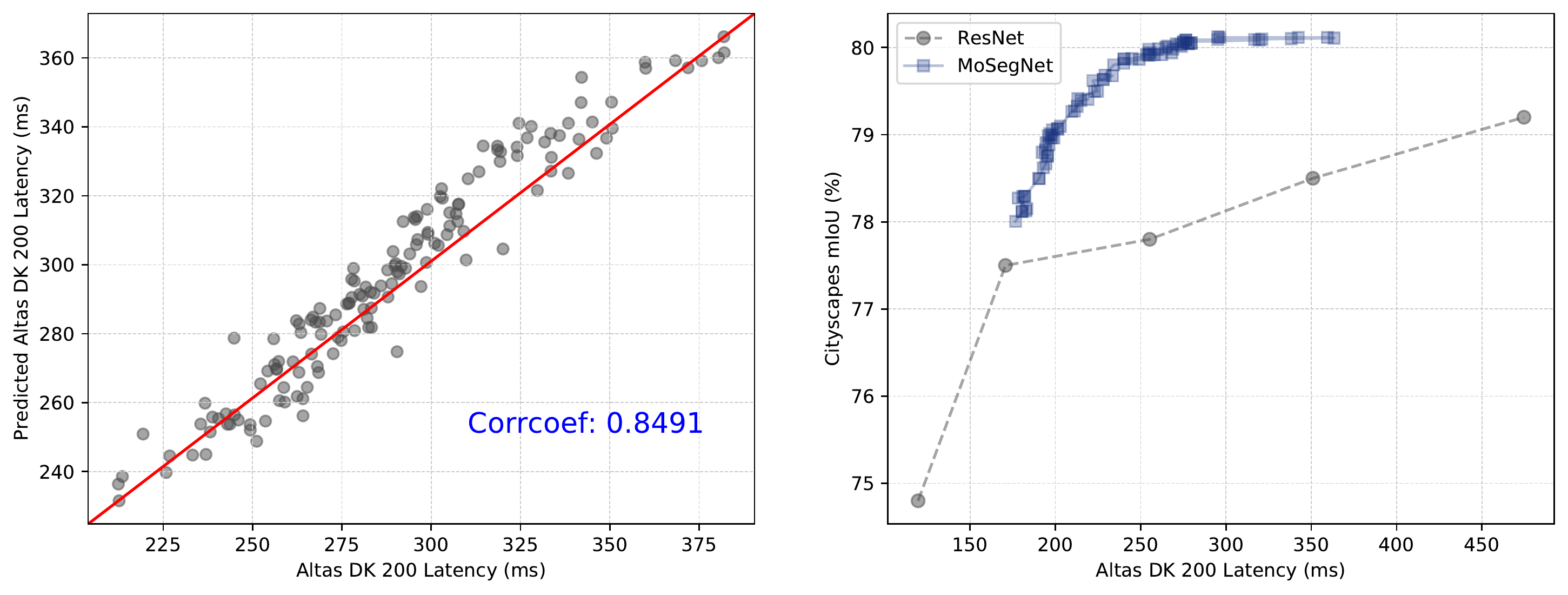}
    \end{subfigure}
\caption{\textbf{(\emph{Left}) Performance of the constructed latency look-up table} of the Huawei Atlas 200 DK. The Kendall correlation coefficient $\tau$ is provided along with the fitted linear model (red line). \textbf{(\emph{Right}) Latency-Accuracy trade-off comparison} between the obtained models and ResNets. Models with lower latency and higher accuracy are preferred (top-left corner).\label{fig:atlas}}
\end{figure}

A portfolio of prior works have demonstrated the necessity of incorporating hardware feedback in the loop of NAS for designing hardware-dependent architectures. 
However, most existing NAS works that target hardware performance were carried out on a simpler task of image classification, and mainly for mobile devices \cite{cai2018proxylessnas,onceforall}. 
In this work, we target a more challenging and demanding task of semantic segmentation. Using the Huawei Atlas 200 Developer Kit (DK) as an example, we demonstrate the practical utility of the proposed algorithm, \ourmethod{}, for designing hardware-dependent models. 
Powered by the Ascend 310 processor, the Huawei Atlas 200 DK delivers 11 TFLOPs@FP16 with merely eight watts of power consumption, making it capable of handling challenging tasks in resource-constrained deployment environments. 

The experimental setup is identical to the previous experiment described in Section~\ref{sec:implement}, where we use the segmentation accuracy on Cityscapes dataset and the inference speed on the Huawei Atlas 200 DK as the twin objectives. We uniformly sample 2K architectures from our search space to construct a look-up table for predicting latency on the Atlas board. The performance of the fitted look-up table is visualized in Fig.~\ref{fig:atlas} (\emph{Left}). 
The returned models at the end of the evolution are re-trained thoroughly from-scratch. Fig.~\ref{fig:atlas} (\emph{Right}) depicts the experimental results, where we observe that the obtained \ourmodel{}s achieve a substantially better speed-accuracy trade-off compared to the ResNets \cite{resnet}. 
More details and visualizations of the obtained architectures are available in the supplementary materials.

\begin{table}[!hbt]
\centering
\caption{Computational efficiency comparison among \ourmodel{}s optimized for FLOPs and latency on RTX 2080Ti and Huawei Atlas 200 DK.\label{tab:atlas}}
\resizebox{.4\textwidth}{!}{%
\begin{tabular}{@{\hspace{2mm}}lccccc@{\hspace{2mm}}}
\toprule
\multirow{2}{*}{\begin{tabular}[c]{@{}c@{}}Optimized \\ for\end{tabular}} & \multirow{2}{*}{\begin{tabular}[c]{@{}c@{}}Cityscapes \\ mIoU (\%)\end{tabular}} & \multirow{2}{*}{\begin{tabular}[c]{@{}c@{}}Params \\ (M)\end{tabular}} & \multirow{2}{*}{\begin{tabular}[c]{@{}c@{}}FLOPs\\ (G)\end{tabular}} & \multicolumn{2}{c}{Latency (ms)} \\ \cmidrule(l){5-6} 
 &  &  &  & GPU & Atlas \\ \midrule
\multicolumn{1}{l}{FLOPs} & 79.02 & 35.3 & \textbf{103.1} & 28.7 & 267.4 \\
\multicolumn{1}{l}{GPU (2080Ti)} & 79.04 & 30.6 & 106.2 & \textbf{26.3} & 261.4 \\
\multicolumn{1}{l}{Atlas 200 DK} & 79.01 & 30.1 & 106.4 & 29.2 & \textbf{196.7} \\ \bottomrule
\end{tabular}%
}
\end{table}

To quantify the algorithmic contribution of \ourmethod{}, the following ablative experiment has been performed. 
We impose an artificial constraint of mIoU being greater than 79\% on Cityscapes dataset, and then search for architectures that are efficient for FLOPs, GPU and Atlas, respectively. 
As can be observed from the results in Table~\ref{tab:atlas}, in general, different computational metrics or hardware require specialized models in order to be efficient. 
It suggests the competing nature of different efficiency metrics and computational environments, leading to the need of considering them in the optimization process, i.e., multi-objective NAS. Furthermore, the experimental results also validate that our proposed algorithm can be a viable mean to design task-specific neural architectures for the challenging task of semantic segmentation.

\section{Conclusion\label{sec:conclusion}}
This paper considered the problem of automating the designs of efficient neural network architectures for semantic segmentation tasks. 
For this purpose, we introduced \emph{\ourmethod{}}, a method harnessing the concept of surrogate modelling within an evolutionary multi-objective framework. 
With a sequence of modifications made to a multi-layer perceptron, we demonstrated that an indicative predictor can be efficiently learned online with few hundreds of samples. 
In the meantime, another surrogate model, in the form of a look-up table, was learned offline and specific to the search space for predicting inference latency. 
Utilizing the fact that high-fidelity evaluation of latency is significantly cheaper than segmentation accuracy, a customized hierarchical pre-screening criterion was proposed for in-fill selection. 
Experimental evaluation on three semantic segmentation datasets showed that models obtained by \ourmethod{} outperform a wide range of state-of-the-art models in terms of both accuracy and inference speed. 
Finally, we demonstrated the practical utility of \ourmethod{} in designing hardware-specific models on an application to the Huawei Atlas 200 DK. 
To the best of our knowledge, \ourmethod{} presents the first realization of evolutionary multi-objective optimization for automatically designing efficient and custom neural network models for the challenging task of real-time semantic segmentation.



\ifCLASSOPTIONcaptionsoff
  \newpage
\fi

\bibliography{egbib} 
\bibliographystyle{IEEEtran}

\end{document}